%% file: main.tex
\documentclass[nohyperref]{article}

\usepackage{microtype}
\usepackage{graphicx}
\usepackage{booktabs} % for professional tables

\usepackage{xcolor}

\usepackage{hyperref}
\usepackage[accepted]{icml2022}
\usepackage{diagbox}
\usepackage{amsmath}
\usepackage{amssymb}
\usepackage{mathtools}
\usepackage{amsthm}
\usepackage[capitalize,noabbrev]{cleveref}

\theoremstyle{plain}

\theoremstyle{definition}

\theoremstyle{remark}

\usepackage[textsize=tiny]{todonotes}

\usepackage{custom}
\usepackage{caption}
\usepackage{subcaption}
\usepackage{adjustbox}
\usepackage{multirow}

\usepackage{tikz}
\usepackage{tablefootnote}
\usepackage{pgfplots, pgfplotstable}
\usepgfplotslibrary{groupplots}
\usepgfplotslibrary{fillbetween}
\usepgfplotslibrary{statistics}
\usepackage{arydshln}

\makeatletter
\def\adl@drawiv#1#2#3{%
        \hskip.5\tabcolsep
        \xleaders#3{#2.5\@tempdimb #1{1}#2.5\@tempdimb}%
                #2\z@ plus1fil minus1fil\relax
        \hskip.5\tabcolsep}
\newcommand{\cdashlinelr}[1]{%
  \noalign{\vskip\aboverulesep
           \global\let\@dashdrawstore\adl@draw
           \global\let\adl@draw\adl@drawiv}
  \cdashline{#1}
  \noalign{\global\let\adl@draw\@dashdrawstore
           \vskip\belowrulesep}}
\makeatother
\icmltitlerunning{Query-Efficient and Scalable Black-Box Adversarial Attacks on Discrete Sequential Data via Bayesian Optimization}
\begin{document}
\twocolumn[
	\icmltitle{Query-Efficient and Scalable Black-Box Adversarial Attacks on \\ Discrete Sequential Data via Bayesian Optimization}



\begin{icmlauthorlist}
\icmlauthor{Deokjae Lee}{yyy}
\icmlauthor{Seungyong Moon}{yyy}
\icmlauthor{Junhyeok Lee}{yyy}
\icmlauthor{Hyun Oh Song}{yyy}
%\icmlauthor{Firstname5 Lastname5}{yyy}
%\icmlauthor{Firstname6 Lastname6}{sch,yyy,comp}
%\icmlauthor{Firstname7 Lastname7}{comp}
%\icmlauthor{Firstname8 Lastname8}{sch}
%\icmlauthor{Firstname8 Lastname8}{yyy,comp}
\end{icmlauthorlist}

\icmlaffiliation{yyy}{Department of Computer Science and Engineering, Seoul National University, Seoul, Korea}
%\icmlaffiliation{comp}{Company Name, Location, Country}
%\icmlaffiliation{sch}{School of ZZZ, Institute of WWW, Location, Country}

\icmlcorrespondingauthor{Hyun Oh Song}{hyunoh@snu.ac.kr}

\icmlkeywords{Machine Learning, ICML}

\vskip 0.3in
]



\printAffiliationsAndNotice{}  % leave blank if no need to mention equal contribution

\begin{abstract}

We focus on the problem of adversarial attacks against models on discrete sequential data in the black-box setting where the attacker aims to craft adversarial examples with limited query access to the victim model. Existing black-box attacks, mostly based on greedy algorithms, find adversarial examples using pre-computed key positions to perturb, which severely limits the search space and might result in suboptimal solutions. To this end, we propose a query-efficient black-box attack using Bayesian optimization, which dynamically computes important positions using an automatic relevance determination (ARD) categorical kernel. We introduce block decomposition and history subsampling techniques to improve the scalability of Bayesian optimization when an input sequence becomes long. Moreover, we develop a post-optimization algorithm that finds adversarial examples with smaller perturbation size. Experiments on natural language and protein classification tasks demonstrate that our method consistently achieves higher attack success rate with significant reduction in query count and modification rate compared to the previous state-of-the-art methods.

\end{abstract}
%\vspace{-2.0em}
\section{Introduction}
%\vspace{-0.4em}
\label{intro}

In recent years, deep neural networks on discrete sequential data have achieved remarkable success in various domains including natural language processing and protein structure prediction, with the advent of large-scale sequence models such as BERT and XLNet \cite{BERT, XLNet}.
However, these networks have exhibited vulnerability against adversarial examples that are artificially crafted to raise network malfunction by adding perturbations imperceptible to humans \cite{papernot2016crafting, TextFooler}. 
Recent works have focused on developing adversarial attacks in the \emph{black-box} setting, where the adversary can only observe the predicted class probabilities on inputs with a limited number of queries to the network \cite{GA, PWWS}. 
This is a more realistic scenario since, for many commercial systems \cite{google, amazon}, the adversary can only query input sequences and receive their prediction scores with restricted resources such as time and cost.

%\vspace{-0.2em}
While a large body of works has proposed successful black-box attacks in the image domain with continuous attack spaces \cite{ilyas2018black, andriushchenko2020square}, developing a query-efficient black-box attack on discrete sequential data is quite challenging due to the discrete nature of their attack spaces. 
%Some prior work regards adversarial attack as a combinatorial optimization problem and applies evolutionary algorithms to solve the problem, but it has been shown that the method requires a large number of queries. \cite{GA, PSO}. 
Some prior works employ evolutionary algorithms for the attack, but these methods require a large number of queries in practice \cite{GA, PSO}. 
Most of the recent works are based on greedy algorithms which first rank the elements in an input sequence by their importance score and then greedily perturb the elements according to the pre-computed ranking for query efficiency \cite{PWWS,TextFooler,LSH}. %However, these algorithms modify each element at most once, which severely limits the search space and shows a lower attack success rate compared to \citet{PSO} \cite{yoo2020searching}.
However, these algorithms have an inherent limitation in that each location is modified at most once and the search space is severely restricted \cite{yoo2020searching}. %Empirically, these methods show lower attack success rate compared to \citet{PSO}.
%might lead to a suboptimal solution \cite{yoo2020searching}.

%\vspace{-0.2em}
To this end, we propose a \emph{Blockwise Bayesian Attack} (BBA) framework, a query-efficient black-box attack based on \emph{Bayesian Optimization}. 
We first introduce a categorical kernel with automatic relevance determination (ARD), suited for dynamically learning the importance score for each categorical variable in an input sequence based on the query history. 
To make our algorithm scalable to a high-dimensional search space, which occurs when an input sequence is long, we devise block decomposition and history subsampling techniques that successfully improve the query and computation efficiency without compromising the attack success rate. Moreover, we propose a post-optimization algorithm that reduces the perturbation size.

%\vspace{-0.2em}
We validate the effectiveness of BBA in a variety of datasets from different domains, including text classification, textual entailment, and protein classification.
Our extensive experiments on various victim models, ranging from classical LSTM to more recent Transformer-based models \cite{LSTM, BERT}, demonstrate state-of-the-art attack performance in comparison to the recent baseline methods.  
Notably, BBA achieves higher attack success rate with considerably less modification rate and fewer required queries on all experiments we consider.

%\vspace{-0.5em}
\section{Related Works}
\label{rel}

\subsection{Black-Box Attacks on Discrete Sequential Data}
Black-box adversarial attacks on discrete sequential data have been primarily studied in natural language processing (NLP) domain, where an input text is manipulated at word levels by substitution \citep{GA}.
A line of research exploits greedy algorithms for finding adversarial examples, which defines the word replacement order at the initial stage and greedily replaces each word under this order by its synonym chosen from a word substitution method \cite{PWWS, TextFooler, LSH, BAE, BERTAttack}.  
\citet{PWWS} determine the priority of words based on word saliency and construct the synonym sets using WordNet \cite{WordNet}.
\citet{TextFooler} construct the word importance ranking by measuring the prediction change after deleting each word and utilize the word embedding space from \citet{mrkvsic2016counter} to identify the synonym sets.
The follow-up work of \citet{LSH} proposes a query-efficient word ranking algorithm that leverages attention mechanism and locality-sensitive hashing. 
Another research direction is to employ combinatorial optimizations for crafting adversarial examples \cite{GA, PSO}.
%\citet{GA} find synonyms for each word according to the distance in GloVe embedding space \citep{GLOVE} and generate adversarial examples via genetic algorithms.
\citet{GA} generate adversarial examples via genetic algorithms.
%\citet{PSO} propose a word-substitution model based on the semantic labels of words, also referred as to sememes, using HowNet \cite{HowNet} and applies particle swarm optimization to solve the problem. 
\citet{PSO} propose a particle swarm optimization-based attack (PSO) with a word substitution method based on sememes using HowNet \cite{HowNet}.

%\vspace{-0.5em}
\subsection{Bayesian Optimization}
While Bayesian optimization has been proven to be remarkably successful for optimizing black-box functions, its application to high-dimensional spaces is known to be notoriously challenging due to its high \emph{query complexity}.
There has been a large body of research that improves the query efficiency of high-dimensional Bayesian optimization.
One major approach is to reduce the effective dimensionality of the objective function using a sparsity-inducing prior for the scale parameters in the kernel \citep{COMBO, SAAS}. 
Several methods address the problem by assuming an additive structure of the objective function and decomposing it into a sum of functions in lower-dimensional disjoint subspaces \citep{kandasamy2015high, wang2018batched}. 
Additionally, a line of works proposes methods that perform multiple evaluation queries in parallel, also referred to as batched Bayesian optimization, to further accelerate the optimization \citep{azimi2010batch, HDBBO}.
% \citet{HDBBO} assign a DPP prior to batch selection to consider the diversity of the batch. 

Another challenge in Bayesian optimization with Gaussian processes (GPs) is its high \emph{computational complexity} of fitting surrogate models on the evaluation history.
A common approach to this problem is to use a subset of the history to train GP models \citep{seeger2003fast, SOD}. 
%\citep{csato2002sparse, seeger2003fast, SOD}. 
\citet{seeger2003fast} greedily select a training point from the history that maximizes the information gain.
\citet{SOD} choose a subset of the history using Farthest Point Clustering heuristic \citep{gonzalez1985clustering}. 

Many Bayesian optimization methods have focused on problem domains with continuous variables. Recently, Bayesian optimization on categorical variables has attained growing attention due to its broad potential applications to machine learning.
\citet{baptista2018bayesian} use Bayesian linear regression as surrogate models for black-box functions over combinatorial structures. 
\citet{COMBO} propose a Bayesian optimization method for combinatorial search spaces using GPs with a discrete diffusion kernel. 

%\citet{CoCaBO} introduce a GP kernel suitable to capture interactions between multiple continuous and categorical variables.

%\vspace{-0.5em}
\subsection{Adversarial Attacks via Bayesian Optimization}
Several works have proposed query-efficient adversarial attacks using Bayesian optimization in image and graph domains, but its applicability to discrete sequential data has not yet been explored.
\citet{Kolter, BayesOpt} leverage Bayesian optimization to attack image classifiers in a low query regime. 
\citet{Kolter} introduce a noise upsampling technique to reduce the input dimensions of image spaces for the scalability of Bayesian optimization.
A concurrent work of \citet{BayesOpt} proposes a new upsampling method, whose resize factor is automatically determined by the Bayesian model selection technique, and adopts an additive GP as a surrogate model to further reduce the dimensionality.
Recently, \citet{GRABNEL} propose a query-efficient attack algorithm against graph classification models using Bayesian optimization with a sparse Bayesian linear regression surrogate. 
While these Bayesian optimization-based methods find adversarial examples with any perturbation size
below a pre-defined threshold, we further consider minimizing the perturbation size, following the practice in the prior works in NLP \citep{PWWS,TextFooler,PSO,LSH}.

%\red{Unlike Bayesian optimization-based attack methods in other domains that search adversarial examples on the attack space with the pre-determined constraint on the perturbation size \citep{Kolter, BayesOpt, GRABNEL}, we search adversarial examples on the attack space without constraint but minimizing perturbation size.}
%{\color{green!80!black!100}  in other domains search adversarial examples on the attack space with the pre-determined constraint on the perturbation size \citep{Kolter, BayesOpt, GRABNEL}. In contrast, we search adversarial examples on the attack space without constraint but consider perturbation size as a metric to minimize, following standard practice in the prior works in NLP }

%\vspace{-0.6em}
\section{Preliminaries}
%\vspace{-0.1em}
\subsection{Problem Formulation}
To start, we introduce the definition of adversarial attacks on discrete sequential data.
Suppose we are given a target classifier $f_\theta: \mathcal{X}^l \to \mathbb{R}^{|\mathcal{Y}|}$, which takes an input sequence of $l$ elements $s = [w_0, \ldots, w_{l-1}] \in \mathcal{X}^l$ and outputs a logit vector used to predict its ground-truth label $y \in \mathcal{Y}$. 
For NLP tasks, $s$ is a text consisting of words $w_i$ from a dictionary $\mathcal{X}$.
Our objective is to craft an adversarial sequence $s_\text{adv}$ that misleads $f_\theta$ to produce an incorrect prediction by replacing as few elements in the input sequence $s$ as possible.
Formally, this can be written as the following optimization problem:
\begin{align}
	\label{maineq}
	&\minimize_{s'\in \mathcal{X}^l} ~ d(s,s')\nonumber\\
	&~ \mathrm{subject~to}~~\mathcal{L}(f_\theta(s'), y) \ge 0,
\end{align}
where $d$ is a distance metric that quantifies the amount of perturbation between two sequences (\eg, Hamming distance) and $\mathcal{L}(f_\theta(s), y) \triangleq \max_{y'\in\mathcal{Y},y'\neq y} f_\theta(s)_{y'} - f_\theta(s)_{y}$ denotes the attack criterion. In this paper, we consider the score-based black-box attack setting, where an adversary has access to the model prediction logits with a limited query budget, but not the model configurations such as network architectures and parameters.

To make the adversarial perturbation imperceptible to humans, the modified sequence should be semantically similar to the original sequence and the perturbation size should be sufficiently small \cite{PWWS}.  
However, minimizing only the perturbation size does not always ensure the semantic similarity between the two sequences. 
For example, in the NLP domain, even a single word replacement can completely change the meaning of the original text due to the characteristics of natural languages. 
To address this, we replace elements with ones that are semantically similar to generate an adversarial example, which is a standard practice in the prior works in NLP. 
Concretely, we first define a set of semantically similar candidates $\mathcal{C}(w_i) \subseteq \mathcal{X}$ for each $i$-th element $w_i$ in the original sequence.
In the NLP domain, this can be found by existing word substitution methods \citep{PWWS, TextFooler, PSO}. 
Then, we find an adversarial sequence in their product space $\prod_{i=0}^{l-1} \mathcal{C}(w_i)\subseteq \mathcal{X}^l$. 

We emphasize that the greedy-based attack methods have the restricted search spaces of size $\sum_{i=0}^{l-1} |\mathcal{C}(w_i)|-l+1$. In contrast, our search space is of cardinality $| \prod_{i=0}^{l-1} \mathcal{C}(w_i) |$, which is always larger than the greedy methods.
% Finally, we choose Hamming distance as the distance metric $d$ and minimize it to reduce the perturbation size.

\subsection{Bayesian Optimization}
\label{main:bo}
Bayesian optimization is one of the most powerful approaches for maximizing a black-box function $g: A \to \mathbb{R}$ \cite{snoek2012practical, frazier2018tutorial}.
It constructs a probabilistic model that approximates the true function $g$, also referred to as a surrogate model, which can be evaluated relatively cheaply.
The surrogate model assigns a prior distribution to $g$ and updates the prior with the evaluation history to get a posterior distribution that better approximates $g$. 
Gaussian processes (GPs) are common choices for the surrogate model due to their flexibility and theoretical properties \cite{osborne2009gaussian}. 
\normalsize A GP prior assumes that the values of $g$ on any finite collection of points $X\subseteq A$ are normally distributed, \ie, $g(X)\sim \mathcal{N}(\mu(X), K(X,X)+\sigma_n^2 I)$, where $\mu: A \to \mathbb{R}$ and $K: A \times A \to \mathbb{R}$ are the mean and kernel functions, respectively, and $\sigma_n^2$ is the noise variance. 
Given the evaluation history $\mathcal{D} = \{(\hat{x}_j, \hat{y}_j=g(\hat{x}_j))\}_{j=0}^{n-1}$, the posterior distribution of $g$ on a finite candidate points $X$ can also be expressed as a Gaussian distribution with the predictive mean and variance as follows:
\begin{align*}
	&\mathrm{E}[g(X) \mid X,\mathcal{D}] \\
	&~~~~= K(X,\hat{X}) [ K(\hat{X},\hat{X})+\sigma_n^2 I ]^{-1} (\hat{Y}-\mu(\hat{X}))+\mu(X) \\
	&\mathrm{Var}[g(X) \mid X, \mathcal{D}] \\
	&~~~~= K(X,X) - K(X,\hat{X}) [ K(\hat{X},\hat{X})+\sigma_n^2 I ]^{-1}K(\hat{X},X),
\end{align*}
where $\hat{X}$ and $\hat{Y}$ are the concatenations of $\hat{x}_j$'s and $\hat{y}_j$'s, respectively.

Based on the current posterior distribution, an acquisition function quantifies the utility of querying $g$ at each point for the purpose of finding the maximizer. 
Bayesian optimization proceeds by maximizing the acquisition function to determine the next point $x_n$ to evaluate and updating the posterior distribution with the new evaluation history $\mathcal{D} \cup \{ (\hat{x}_n, g(\hat{x}_n)) \}$. 
After a fixed number of function evaluations, the point evaluated with the largest $g(x)$ is returned as the solution.
\begin{figure*}[h]
	\centering
	\includegraphics[width=0.9\textwidth]{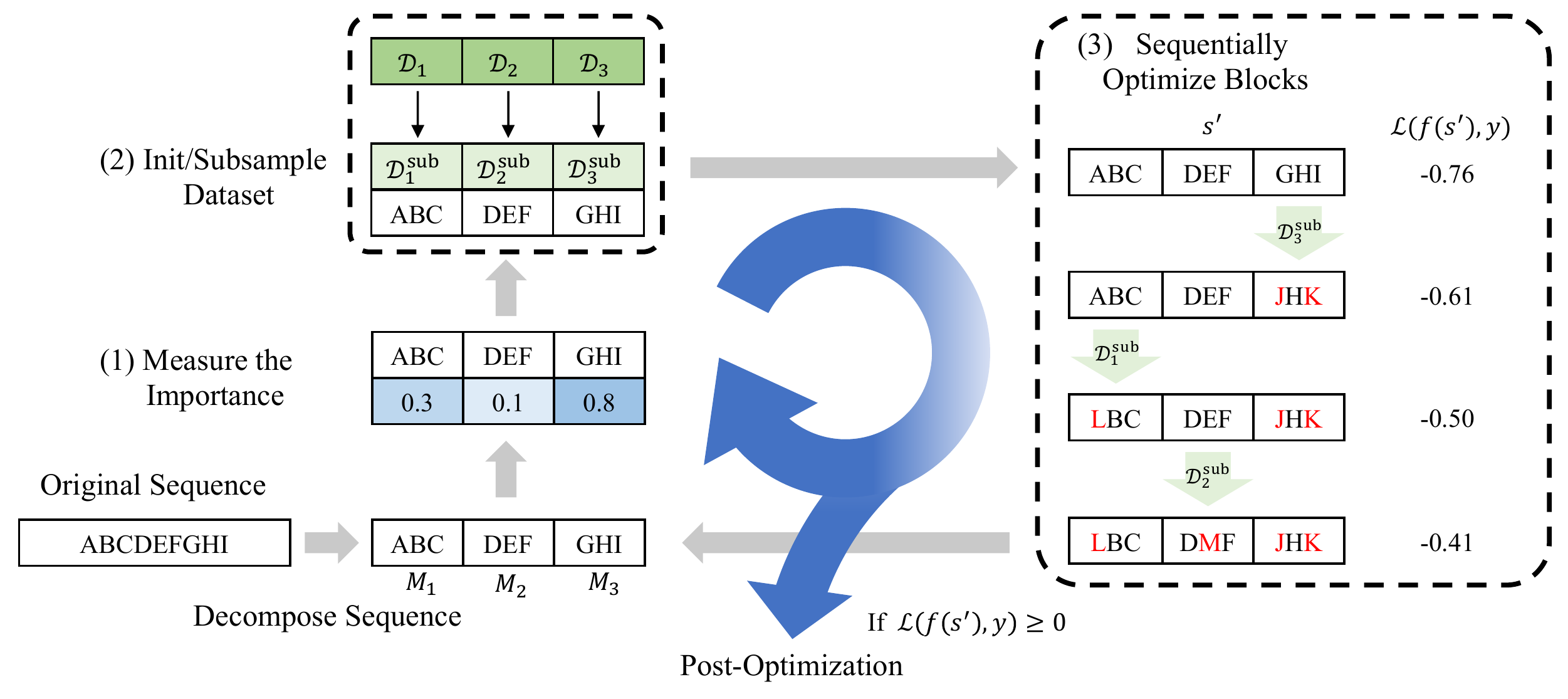}
	\caption{The overall process of BBA. A green arrow with a dataset $\mathcal{D}_k^\text{sub}$ denotes the Bayesian optimization step for the block $M_k$ using $D_k^\text{sub}$ as the initial dataset.}
	\label{fig:proc}
\end{figure*}
%\vspace{-0.5em}
% 4. Methods
\section{Methods}
\label{method}
In this section, we introduce the proposed \emph{Blockwise Bayesian Attack} (BBA) framework. Instead of optimizing \cref{maineq} directly, we divide the optimization into two steps. 
First, we conduct Bayesian optimization to maximize the black-box function $\mathcal{L}(f_\theta(\cdot), y)$ on the attack space 
$\mathcal{S} \triangleq \prod_{i=0}^{l-1} \mathcal{C}(w_i)$ until finding an adversarial sequence $s_\text{adv}$, which is a feasible solution of \cref{maineq}. This step can be formulated as
\begin{align}
	\label{subeq}
	\mathop{\mathrm{maximize}}_{s' \in \mathcal{S}} ~ \mathcal{L}(f_\theta(s'), y).
\end{align}
Second, after finding a valid adversarial sequence $s_\text{adv}$ that satisfies the attack criterion $\mathcal{L}(f_\theta(s_\text{adv}), y) \ge 0$, 
we seek to reduce the Hamming distance of the perturbed sequence from the original input while maintaining the constant feasibility.

Note that \cref{subeq} is a high-dimensional Bayesian optimization problem on combinatorial search space, especially for datasets consisting of long sequences.
However, the number of queries required to obtain good coverage of the input space, which is necessary to find the optimal solution, increases exponentially with respect to the input dimensions due to the curse of dimensionality \cite{shahriari2015taking}.
This high \emph{query complexity} is prohibitive for query-efficient adversarial attacks. Furthermore, even in a low-dimensional space, the high \emph{computational complexity} of training GP models in Bayesian optimization can drastically slow down the runtime of the algorithm as the evaluation history becomes larger. Fitting the GP model requires the matrix inversion of the covariance matrix $K(\hat{X},\hat{X})$, whose computational complexity is $\mathcal{O}(n^3)$, where $n$ is the number of evaluations so far.

To this end, we first introduce the surrogate model and the parameter fitting method which are suitable for our high-dimensional combinatorial search space. 
Next, we propose two techniques to deal with the scalability issues that arise from the high query and computational complexity of Bayesian optimization. 
Lastly, we introduce a post-optimization technique that effectively minimizes the perturbation size of an adversarial sequence.

\subsection{Surrogate Model and GP Parameter Fitting}
Choosing an appropriate kernel that captures the structure of the high-dimensional combinatorial search space is the key to the success of our GP-based surrogate model. 
We use a categorical kernel\footnote{\url{https://botorch.org/api/_modules/botorch/models/kernels/categorical.html}} with automatic relevance determination (ARD) to automatically determine the degree to which each input dimension is important \citep{mackay1992bayesian}. 
The kernel has the following form: 
\begin{align*}
	K^\text{cate}(s^{(1)}, s^{(2)}) = \sigma_f^2 \prod_{i=0}^{l-1} \exp \left( - \frac{\mathbf{1}[w_i^{(1)} \neq w_i^{(2)}]}{\beta_i} \right),
\end{align*}
where $\sigma_f^2$ is a signal variance, $\beta_i$ is a length-scale parameter corresponding to the relevance of $i$-th element position. 
This implies that the kernel regards a sequence pair sharing a larger number of elements as a more similar pair.
The GP parameter $\beta_i$ is estimated by maximizing the posterior probability of the evaluation history under a prior using the gradient descent with Adam optimizer \cite{Adam}. 
More details can be found in \cref{app:GP}.

\subsection{Techniques for Scalability}
\label{scalability}
To achieve a scalable Bayesian optimization algorithm, we decompose an input sequence into disjoint blocks of element positions and optimize each block in a sequential fashion for several iterations using data subsampled from the evaluation history corresponding to the block.

\subsubsection{Block Decomposition}
We divide an input sequence of length $l$ into $\lceil l/m \rceil$ disjoint blocks of length $m$.
Each $k$-th block $M_k$ consists of consecutive indices $[km, \ldots, (k+1)m-1]$. 
We sequentially optimize each block for $R$ iterations, rather than updating all element positions concurrently. 
For each iteration, we set the maximum query budget to $N_k$ when optimizing the block $M_k$.
While the dimension of the attack space $\prod_{i=0}^{l-1} \mathcal{C}(w_i)$ grows exponentially as $l$ increases, the block decomposition makes the dimension of the search space of each Bayesian optimization step independent of $l$ and upper bounded by $(C_\text{max})^m$, where $C_\text{max}\triangleq\max_{i\in[l]}|\mathcal{C}_i|$ is the size of the largest synonym set.

%For the first iteration, we optimize the blocks in order of increasing block index $k$.
At the start of each iteration, we assign an importance score to each block, which measures how much each block contributes to the objective function value. Then, we sequentially optimize blocks in order of highest importance score for query efficiency. For the first iteration, we set the importance score of each block to the change in the objective function value after deleting the block. For the remaining iterations, we reassign the importance score to each block $M_k$ by summing the inverses of the length-scale parameters that correspond to the element positions in $M_k$, \ie, $\sum_{i\in M_k}1/\beta_i$.
%This importance score can give a sensitivity measure of how much each block contributes to the objective function value.

%For the first iteration, we determine the importance of each block as the prediction difference after deleting each block. Then, we optimize the blocks in order of decreasing block importance.
%\textcolor{blue}{After} each iteration ends, we assign an importance score to each block $M_k$ by summing the values of the length-scale parameters that correspond to the word positions in $M_k$, \ie, $\sum_{i\in M_k}\beta_i$.
%This importance score can give a sensitivity measure of how much each block contributes to the objective function value.
%zThen, we optimize the blocks in order of highest importance score for query efficiency.
\begin{algorithm}[t]
	\caption{SoD$(\mathcal{D},N)$, Subset of Data method}
	\label{alg:sod}
	\begin{algorithmic}[1]
		\STATE {\bfseries Input:} The evaluation history $\mathcal{D}$, the size of subsamples $N$. 
		\IF{$|\mathcal{D}| < N$}
			\STATE {\bfseries Return} $\mathcal{D}$.
		\ENDIF
		\STATE Initialize the dataset $\mathcal{D}_\text{sub} \leftarrow \{s_0\}$ where $s_0$ is randomly sampled from $\mathcal{D}$.
		\WHILE {$|\mathcal{D}_\text{sub}| < N$}
			\STATE Select the farthest sequence. \\
			$s_\text{far} \leftarrow \argmax_{s\in \mathcal{D}\setminus\mathcal{D}_\text{sub}} [\min_{s'\in\mathcal{D}_\text{sub}}d(s,s')]$
			\STATE Update the dataset $\mathcal{D}_\text{sub}\leftarrow \mathcal{D}_\text{sub} \cup \{s_\text{far}\}$.
		\ENDWHILE
		\STATE {\bfseries Return} $\mathcal{D}_\text{sub}$.
	\end{algorithmic}
\end{algorithm}

\begin{algorithm}[ t]
	\caption{PostOpt$(s, s_\text{adv},\mathcal{D}_\text{sub},N_\text{post},N_b)$}%, Post-optimization process}
	\label{alg:postopt}
	\begin{algorithmic}[1]
		\STATE {\bfseries Input:} The original sequence $s$, an adversarial sequence $s_\text{adv}$, the evaluation dataset $\mathcal{D}_\text{sub}$ subsampled from the evaluation history, the query budget $N_\text{post}$, and the batch size $N_b$. 
		%\STATE Fit parameters $\{\beta_i\}$, $\sigma_f^2$ to maximize posterior probability distribution $p(\{\beta_i\},\sigma_f^2\mid\mathcal{D}^\text{sub})$.
		%\FOR{$k=0$ {\bfseries to} $l/m-1$}
			\STATE Initialize $N_r \leftarrow N_\text{post}$.
			\WHILE {$N_r > 0$}
				%\STATE Fit parameters $\{\beta_i\}_{i=0}^{l-1}$, $\sigma_f^2$ to maximize posterior probability distribution $p(\{\beta_i\}_{i=0}^{l-1},\sigma_f^2\mid\mathcal{D}_{j[k]}^\text{selected})$.
				%\STATE Fit parameters $\{\beta_i\}$, $\sigma_f^2$ to maximize the posterior probability distribution $p(\{\beta_i\},\sigma_f^2\mid\mathcal{D}_\text{sub})$.
				\STATE Fit GP parameters to maximize the posterior probability distribution on $\mathcal{D}_\text{sub}$.
				\STATE Select a batch $B$ of the size $\min(N_b, N_r)$ from $\mathcal{B}_H(s,d_H(s,s_\text{adv})-1) \cap \mathcal{B}_H(s_\text{adv},r)$ according to the acquisition function and the DPP.
				\STATE Evaluate the batch $\mathcal{D}_\text{batch} = \{(s',\mathcal{L}(f_\theta(s'), y)\}_{s'\in B}$.
				\STATE Update the dataset $\mathcal{D}_\text{sub}\leftarrow \mathcal{D}_\text{sub}\cup \mathcal{D}_\text{batch}$.
				\STATE $N_r \leftarrow N_r - |\mathcal{D}_\text{batch}|$.
				\IF {$B$ has an adversarial sequence}
					\STATE Update $s_\text{adv}$ to the best adversarial sequence in $B$.
					\STATE $N_r \leftarrow N_\text{post}$.
				\ENDIF
			\ENDWHILE
		%\ENDFOR
		\STATE {\bfseries Return} $s_\text{adv}$.
	\end{algorithmic}
\end{algorithm}
\subsubsection{History Subsampling}
Here, we propose a data subsampling strategy suitable for our block decomposition method. 
When we optimize a block $M_k$, only the elements in $M_k$ are updated while the remaining elements are unchanged.
Thus, in terms of the block $M_k$, all sequences evaluated during the optimization steps for blocks other than $M_k$ share the same elements, which do not provide any information on how much $M_k$ affects the objective function value.
To avoid this redundancy, we consider utilizing only the sequences collected from the previous optimization steps for $M_k$ as the evaluation history, denoted by $\mathcal{D}_k$, when optimizing $M_k$.

On top of the strategy above, we further reduce the computational complexity of Bayesian optimization by subsampling a dataset from the evaluation history and training the GP surrogate model with the reduced dataset.
We adopt the Subset of Data (SoD) method with Farthest Point Clustering (FPC) \cite{SOD}, a simple and efficient subsampling method widely used in the GP literature.
Concretely, we randomly sample an initial sequence from the evaluation history and sequentially select the farthest sequence that maximizes the Hamming distance to the nearest of all sequences picked so far.
The overall procedure is shown in \cref{alg:sod}.
When optimizing a block $M_k$ at each iteration, we select a subset $\mathcal{D}_k^\text{sub}$ from the evaluation history $\mathcal{D}_k$ via the subsampling algorithm above and proceed with the Bayesian optimization step for $M_k$ using $\mathcal{D}_k^\text{sub}$ as the initial dataset for the GP model training. 

\label{main:ra}
Here, we simply set the initial subset size to $N_k$, which is the same as the maximum query budget when optimizing the block $M_k$.
Thus, the size of the dataset $\mathcal{D}_k^\text{sub}$ during a single block optimization step is upper bounded by $\mathcal{O}(N_k)$. 
Therefore, we can write the complexity of the GP model fitting step when optimizing a block by $\mathcal{O}((\max_k{N_k})^3)$, which is independent of the total number of evaluations, $n$. More details containing the runtime analysis of the overall process can be found in \cref{app:exp,app:RA}.
%\blue{To achieve the complexity independent of query count $n$, we set the size of subsamples to $N_k$, which is the query budget when optimizing the block $M_k$, to make the size of the dataset $\mathcal{D}_k^\text{sub}$ during the optimization upper bounded by $\mathcal{O}(N_k)$. Accordingly, we can write the complexity of the GP model fitting by $\mathcal{O}(N_k^3)$, which is independent of $n$. All the implementation details containing runtime analysis of the overall process can be found in \cref{app:ID}}.

%For a fixed number of queries, the smaller block size $m$ leads to a smaller number of queries for each block, which makes the optimization algorithm run faster in practice.
%However, if $m$ is small, the algorithm cannot capture the relationship between the elements, which might result in a suboptimal solution. 
%If $m$ is larger than the sentence length $l$, our block decomposition method treats the whole sequence as a single block and the algorithm operates the same as without block decomposition. 

\begin{table*}[hbt!]
    \centering 
\caption{Attack results for XLNet-base, BERT-base, and LSTM models on sentence-level classification datasets.}
\begin{subtable}[h!]{0.655 \columnwidth}
	\label{tab:main1}
	\caption{WordNet}
\centering
\begin{adjustbox}{max width=\columnwidth}
\begin{tabular}{cccccc}
    \toprule
Dataset&Model&Method & ASR (\%)& MR (\%)& Qrs \\
\midrule
AG&BERT-base& PWWS& 57.1& 18.3&   367\\
&    & BBA& \textbf{77.4}& \textbf{17.8}&   \textbf{217}\\ %% 50
\cmidrule{2-6}
&LSTM& PWWS& 78.3& 16.4&   336\\
&    & BBA& \textbf{83.2}& \textbf{15.4}&   \textbf{190}\\ %% 50
\midrule
MR&XLNet-base& PWWS& 83.9& \textbf{14.4}&   143\\
&    & BBA& \textbf{87.8}& \textbf{14.4}&    \textbf{77}\\ %% 50
\cmidrule{2-6}
&BERT-base& PWWS& 82.0& 15.0&   143\\
&    & BBA& \textbf{88.3}& \textbf{14.6}&    \textbf{94}\\ %% 100
\cmidrule{2-6}
&LSTM& PWWS& \textbf{94.2}& 13.3&   132\\
&    & BBA& \textbf{94.2}& \textbf{13.0}&    \textbf{67}\\ %% 50
\bottomrule
\end{tabular}
\end{adjustbox}
\end{subtable}
    \begin{subtable}[h!]{0.655 \columnwidth}
\caption{Embedding}
\centering
\begin{adjustbox}{max width=\columnwidth}
\begin{tabular}{cccccc}
    \toprule
Dataset&Model&Method & ASR (\%)& MR (\%)& Qrs \\

\midrule
AG&BERT-base&   TF& 84.7& 24.9&   346\\
&    & BBA& \textbf{96.0}& \textbf{18.9}&   \textbf{154}\\ %% 20
\cmidrule{2-6}
&LSTM&   TF& 94.9& 17.3&   228\\
&    & BBA& \textbf{98.5}& \textbf{16.6}&   \textbf{142}\\ %% 20
\midrule
MR&XLNet-base&   TF& 95.0& 18.0&   101\\
&    & BBA& \textbf{96.3}& \textbf{16.2}&    \textbf{68}\\ %% 20
\cmidrule{2-6}
&BERT-base&   TF& 89.2& 20.0&   115\\
&    & BBA& \textbf{95.7}& \textbf{16.9}&    \textbf{67}\\ %% 20
\cmidrule{2-6}
&LSTM&   TF& \textbf{98.2}& 13.6&    72\\
&    & BBA& \textbf{98.2}& \textbf{13.1}&    \textbf{54}\\ %% 20
\bottomrule
\end{tabular}
\end{adjustbox}
\end{subtable}
\begin{subtable}[h!]{0.682 \columnwidth}
    \caption{HowNet}
    \centering
    \begin{adjustbox}{max width=\columnwidth}
    \begin{tabular}{cccccc}
        \toprule
Dataset&Model&Method & ASR (\%)& MR (\%)& Qrs \\
\midrule
AG&BERT-base&  PSO& 67.2& 21.2& 65860\\
&    &BBA& \textbf{70.8}& \textbf{15.5}&  \textbf{5176}\\ %% 100
\cmidrule{2-6}
&LSTM&  PSO& 71.0& 19.7& 44956\\
&    & BBA& \textbf{71.9}& \textbf{13.7}&  \textbf{3278}\\ %% 100
\midrule
MR&XLNet-base&  PSO& \textbf{91.3}& 18.6&  4504\\
&    & BBA& \textbf{91.3}& \textbf{11.7}&   \textbf{321}\\ %% 100
\cmidrule{2-6}
&BERT-base&  PSO& \textbf{90.9}& 17.3&  6299\\
&    & BBA& \textbf{90.9}& \textbf{12.4}&   \textbf{403}\\ %% 100
\cmidrule{2-6}
&LSTM&  PSO& \textbf{94.4}& 15.3&  2030\\
&    & BBA& \textbf{94.4}& \textbf{11.2}&   \textbf{138}\\ %% 100
\bottomrule
    \end{tabular}
    \end{adjustbox}
    \end{subtable}
\end{table*}
\begin{table*}[hbt!]
    \centering
\caption{Attack results for BERT-base models on document-level classification datasets.}
\begin{subtable}[h!]{0.665\columnwidth}
	\label{tab:main2}
	\caption{WordNet}
\centering
\begin{adjustbox}{max width=\columnwidth}
\begin{tabular}{cccccc}
    \toprule
Dataset&Method & ASR (\%)& MR (\%)& Qrs \\
\midrule
IMDB& PWWS& 97.6& 4.5&  1672\\
   & BBA& \textbf{99.6}& \textbf{4.1}&   \textbf{449}\\ % 100
\cmidrule{2-5}
    &  LSH& 96.3& 5.3&   557\\
    & BBA& \textbf{98.9}& \textbf{4.8}&   \textbf{372}\\ % 50
\cmidrule{1-5}
Yelp& PWWS& 94.3& 7.6&  1036\\
    & BBA& \textbf{99.2}& \textbf{7.4}&   \textbf{486}\\ % 200
\cmidrule{2-5}
    &  LSH& 92.6& 9.5&   389\\
    & BBA& \textbf{98.8}& \textbf{8.8}&   \textbf{271}\\ % 50
	\bottomrule 
\end{tabular}
\end{adjustbox}
\end{subtable}
\begin{subtable}[h!]{0.65\columnwidth}
	\caption{Embedding}
	\centering
	\begin{adjustbox}{max width=\columnwidth}
	\begin{tabular}{cccccc}
		\toprule
Dataset&Method & ASR (\%)& MR (\%)& Qrs \\
\midrule
IMDB&   TF& 99.1& 8.6&   712\\
    & BBA& \textbf{99.6}& \textbf{6.1}&   \textbf{339}\\ % 20
\cmidrule{2-5}
    &  LSH& 98.5& 5.0&   770\\
    & BBA& \textbf{99.8}& \textbf{4.9}&   \textbf{413}\\ % 50
\cmidrule{1-5}
Yelp&   TF& 93.5& 11.1&   461\\
    & BBA& \textbf{99.8}& \textbf{9.6}&   \textbf{319}\\ % 20
\cmidrule{2-5}
        &  LSH& 94.7& 8.9&   550\\
        & BBA& \textbf{99.8}& \textbf{8.6}&   \textbf{403}\\ % 100
	\bottomrule
\end{tabular}
\end{adjustbox}
\end{subtable}
\begin{subtable}[h!]{0.695\columnwidth}
	\caption{HowNet}
	\centering
	\begin{adjustbox}{max width=\columnwidth}
	\begin{tabular}{cccccc}
		\toprule
Dataset&Method & ASR (\%)& MR (\%)& Qrs \\
\midrule
IMDB&  PSO& \textbf{100.0}& 3.8& 113343\\
    & BBA& \textbf{100.0}& \textbf{3.3}&   \textbf{352}\\ % 50
\cmidrule{2-5}
    &  LSH& 98.7& 3.2&   640\\
    & BBA& \textbf{99.8}& \textbf{3.0}&   \textbf{411}\\ % 100
\cmidrule{1-5}
Yelp&  PSO& \textbf{98.8}& 10.6& 86611\\
    & BBA& \textbf{98.8}& \textbf{8.2}&   \textbf{283}\\ % 50
\cmidrule{2-5}
    &  LSH& 93.9& 8.0&   533\\
    & BBA& \textbf{98.2}& \textbf{7.4}&   \textbf{353}\\ % 100
\bottomrule
\end{tabular}
\end{adjustbox}
\end{subtable}
\end{table*}

\subsubsection{Acquisition Maximization Considering Batch Diversity via Determinantal Point Process}
We utilize expected improvement as the acquisition function, which is defined as $\mathrm{EI}(x) = \mathrm{E}[\max(g(s)-g^*_\mathcal{D},0)]$,
where $g^*_\mathcal{D} = \max_{\hat{y}\in \hat{Y}} \hat{y}$ is the largest value evaluated so far.

To further enhance the runtime of the Bayesian optimization algorithm, we evaluate a batch of sequences parallelly in a single round, following the practice in \citet{HDBBO}. 
We sample an evaluation batch $B$ via a Determinantal Point Process (DPP), which promotes batch diversity by maximizing the determinant of its posterior variance matrix $\mathrm{Var}(g(B) \mid \mathcal{D})$ \citep{kulesza2012determinantal}. 
Concretely, we first select sequences with top-$T$ acquisition values in the 1-Hamming distance ball $\mathcal{B}_H(s^*_\mathcal{D},1)$ of the best sequence $s^*_\mathcal{D}$ evaluated so far.
Then, we greedily choose $N_b$ sequences among the top-$T$ sequences that maximize the determinant. More details can be found in \cref{app:exp}.

\subsection{Post-Optimization for Perturbation Reduction}
\label{postopt}
Since we do not consider the perturbation size during the first step of BBA, we conduct a post-optimization step to reduce the perturbation size.
To this end, we optimize for a sequence near the current adversarial sequence $s_\text{adv}$ that stays adversarial and has a smaller perturbation than $s_\text{adv}$.
To achieve this, we search an adversarial sequence in a reduced search space $\mathcal{B}_H(s,d_H(s,s_\text{adv})-1) \cap \mathcal{B}_H(s_\text{adv},r)$, where $r$ controls the exploration size. 
We also conduct Bayesian optimization for the post-optimization step. 
We leverage the evaluation history collected during the first step of BBA and subsample an initial dataset for the GP model training from the history.
If we find a new adversarial sequence during this step, we replace the current adversarial sequence with the new sequence and 
repeat the step above until we cannot find a new adversarial sequence using the query budget $N_\text{post}$ after the most recent update.
The overall post-optimization procedure is summarized in \cref{alg:postopt}.

\cref{fig:proc} illustrates the overall process of BBA.
%The process halts when it reaches the maximum number of iterations or find sequence satisfying attack criterion. 
Please refer to \cref{alg:main} in \cref{app:mainalg} for the more detailed overall algorithm of BBA.

\section{Experiments}
We evaluate the performance of BBA on text classification, textual entailment, and protein classification tasks. We first provide a brief description of the datasets, victim models, and baseline methods used in the experiments. Then, we report the performance of BBA compared to the baselines. Our implementation is available at \url{https://github.com/snu-mllab/DiscreteBlockBayesAttack}.
\input{fig2.tex}

\subsection{Datasets and Victim Models}
To demonstrate the wide applicability and effectiveness of BBA, we conduct experiments on various datasets in the NLP and protein domain. In the NLP domain, we use sentence-level text classification datasets (AG's News, Movie Review), document-level classification datasets (IMDB, Yelp), and textual entailment datasets (MNLI, QNLI) \citep{Yelp_and_AG,MR,IMDB,MNLI,QNLI}. 
In the protein domain, we use an enzyme classification dataset (EC) with 3-level hierarchical multi-labels \citep{EC50}. 
%In protein domain, we use a multi-label enzyme classification dataset (EC50) with three label levels, 0, 1, and 2 \citep{EC50}. 
Note that a protein is a sequence of amino acids, each of which is a discrete categorical variable. 

We consider multiple types of victim models to attack, including bi-directional word LSTM, ASGD Weight-Dropped LSTM, fine-tuned BERT-base and BERT-large, and fine-tuned XLNet-base and XLNet-large \cite{LSTM, AWD-LSTM, BERT, XLNet}. More details on datasets and victim models can be found in \cref{app:datasets,app:models}, respectively.

\subsection{Baseline Methods}
In the NLP domain, we compare the performance of BBA against the state-of-the-art methods such as PWWS, TextFooler, LSH, BAE, and PSO, the first four of which are greedy-based algorithms \cite{PWWS,TextFooler, LSH, BAE, PSO}.
Note that PWWS, TextFooler, BAE, and PSO have different attack search spaces since they utilize different word substitution methods (WordNet, Embedding, BERT masked language model, and HowNet, respectively) \cite{WordNet,mrkvsic2016counter,HowNet}.
For a fair comparison, we follow the practice in \citet{LSH} and compare BBA against each baseline individually under the same attack setting (\eg, word substitution method, query budget) as used in the baseline.
% follow the practice in \citet{LSH} and compares against PWWS, TextFooler, and PSO in the same attack setting (e.g., word substitution method). 
%In the NLP domain, we compare the performance of BBA against the current state-of-the-art methods such as PWWS, TextFooler, LSH, and particle swarm optimization-based attack (PSO), the first three of which are greedy-based algorithms \cite{PWWS,TextFooler, LSH, PSO}.
%We note that PWWS, TextFooler, and PSO use different word substitution methods: WordNet, word embedding, and HowNet, respectively, resulting in different search spaces \cite{WordNet,mrkvsic2016counter,HowNet}.
%On the other hand, LSH can be applied to all three search spaces.
%For a fair comparison, we compare BBA against each baseline in the same attack settings, following the practice in \citet{LSH}. 
We also note that LSH leverages additional attention models, each of which is pre-trained on a different classification dataset. Please refer to \cref{app:spaces} for more details.

For the protein classification task, we compare BBA with TextFooler.
To define its attack space, we exploit the experimental exchangeability of amino acids \citep{ExEx}, which quantifies the mean effect of exchanging one amino acid to a different amino acid on protein activity, as the measure of semantic similarity. 
Then, we define a synonym set for each amino acid by thresholding amino acids with the experimental exchangeability and set the attack space to the product of the synonym sets. 
As in the NLP domain, we compare BBA with the baseline under the same experimental setting as used in the baseline.

\subsection{Attack Performance}

We quantify the attack performance in terms of three main metrics: attack success rate (ASR), modification rate (MR), and the average number of queries (Qrs). 
The attack success rate is defined as the rate of successfully finding misclassified sequences from the original sequences that are correctly classified, which directly measures the effectiveness of the attack method. 
The modification rate is defined as the percentage of modified elements after the attack, averaged over successfully fooled sequences. %sequences being attacked,
This rate is formally written by $\mathrm{E}[d_H(s,s_\text{adv}) / \mathrm{len}(s)]$, which quantifies the distortion of the perturbed sequences from the original.
%\blue{The , which represents the query efficiency of the attack methods, is averaged over the sequences being attacked.} 
The average number of queries, computed over all sequences being attacked, represents the query efficiency of the attack methods.

The main attack results on text classification tasks are summarized in \cref{tab:main1,tab:main2}. 
The results show that BBA significantly outperforms all the baseline methods in all the evaluation metrics for all datasets and victim models we consider. 
%Also, we achieve state-of-the-art performance in all three evaluation metrics on textual entailment tasks, as shown in \cref{tab:main3}. 
\cref{fig:main} shows the cumulative distribution of the number of queries required for the attack methods against a BERT-base model on the Yelp dataset. 
The results show that BBA finds successful adversarial texts using fewer queries than the baseline methods.
More experimental results on other target models (BERT-large, XLNet-large), baseline method (BAE), and datasets (MNLI, QNLI) can be found in  \cref{app:add}. 
%\cref{tab:large,tab:BAE,tab:main3} of \cref{app:add}. 

Moreover, \cref{tab:protein} shows that BBA outperforms the baseline method by a large margin for the protein classification task, which shows the general applicability and effectiveness of BBA on multiple domains. 
\begin{table}[ht]	
	\centering
	\caption{Attack results against AWD-LSTM models on the protein classification dataset EC50 level 0, 1, and 2.}
	\begin{adjustbox}{max width=\columnwidth}
		\begin{tabular}{cccccccccc}
			\toprule
				&&Level 0 &&&Level 1 &&& Level 2&\\
				\cmidrule(lr){2-4} \cmidrule(lr){5-7} \cmidrule(lr){8-10}
			Method	&ASR &MR	&Qrs &ASR &MR	&Qrs &ASR &MR &Qrs\\
			\midrule
			TF & 83.8 & 3.2 & 619 & 85.8 & 3.0 & 584 & 89.6 & 2.5 & 538 \\
			BBA & \textbf{99.8} & \textbf{2.9} & \textbf{285} & \textbf{99.8} & \textbf{2.3} & \textbf{293} & \textbf{100.0} & \textbf{2.0} & \textbf{231}\\
			\bottomrule
		\end{tabular}
	\end{adjustbox}
	\label{tab:protein}
\end{table}

% \blue{In our experiment, the modification rate is averaged over the successfully fooled sequences following practice of the prior works. However, it is unfair since each method averages the modification rate in a different text set.}
 
For a direct comparison with a baseline, one can compute the MR and Qrs over the texts that both BBA and the baseline method are successful on. \Cref{tab:main1anal} shows that BBA outperforms PWWS in MR and Qrs on samples that both methods successfully fooled by a larger margin.\footnote{For BERT-base on AG, PWWS fools 267 texts, BBA fools 363 texts, and both commonly fools 262 texts among 500 texts. For LSTM on AG, PWWS fools 354 texts, BBA fools 376 texts, and both commonly fools 349 texts among 500 texts.}
\begin{table}[ht]
	\caption{Attack results on the AG's News. MR and Qrs of `both success' are averaged over the texts that both PWWS and BBA successfully fooled.}
\centering
\begin{adjustbox}{max width=\columnwidth}
\begin{tabular}{ccccccc}
    \toprule
	&&&&&\multicolumn{2}{c}{Both success}\\
	\cmidrule(lr){6-7}
Model&Method & ASR (\%)& MR (\%)& Qrs &MR (\%) & Qrs  \\
\midrule
BERT-base& PWWS& 57.1& 18.3&   367 & 17.8 & 311 \\
    & BBA& \textbf{77.4}& \textbf{17.8}&   \textbf{217} & \textbf{14.0} & \textbf{154}\\ %% 50
\cmidrule{1-7}
LSTM& PWWS& 78.3& 16.4&   336 & 16.1 & 311\\
    & BBA& \textbf{83.2}& \textbf{15.4}&   \textbf{190} & \textbf{14.4} & \textbf{163}\\ %% 50
\bottomrule
\end{tabular}
\end{adjustbox}
	\label{tab:main1anal}
\end{table}

\subsection{Ablation Studies}

\subsubsection{The Effect of DPP in Batch Update}
To validate the effectiveness of the DPP-based batch update technique, we compare BBA with the greedy-style batch update which chooses the sequences of top-$N_b$ acquisition values for the next evaluations. We do not utilize the post-optimization process to isolate the effect of the batch update.
%We do not utilize other techniques to isolate the effect of the batch update. %\textcolor{blue}{We conduct BBA without query budget, with the max number of iterations of 5}. 
%\cref{tab:abl-batch} shows that batch update technique not only reduces the runtime of the algorithm, but also achieves slight increase in both metrics, which indicates effectiveness of diversified batch update techinique via DPP. 
\cref{tab:abl-batch} shows that the batch update with DPP consistently achieves higher attack success rate using fewer queries compared to the greedy-style batch update. Surprisingly, the batch update with DPP achieves higher attack success rate using fewer queries compared to `without batch update' in AG's News dataset.
\begin{table}[ht]	
	\caption{Attack results of BBA with and without batch update using the WordNet-based word substitution against BERT-base and LSTM models on the sentence-level classification datasets. `Top-$N_b$' denotes the greedy-style batch update that chooses sequences of top-$N_b$ acquisition values.}
	\centering
	\begin{adjustbox}{max width=\columnwidth}
		\begin{tabular}{cccccc}
			\toprule
			&&\multicolumn{2}{c}{BERT-base}&\multicolumn{2}{c}{LSTM}\\
			\cmidrule(lr){3-4}\cmidrule(lr){5-6}
			Dataset& Method &ASR (\%)	&Qrs & ASR (\%) & Qrs\\
			\midrule
			AG& w/o batch& 76.1& 126 & 73.5&  127\\
			\cmidrule{2-6}
			    & w/ batch, Top-$N_b$& 75.9&  133 & 74.3& 127\\
			    & w/ batch, DPP& \textbf{77.4}&  \textbf{124}& \textbf{83.2}&   \textbf{86}\\
				\midrule
			MR& w/o batch& 88.5&  26& 93.9&    18\\
			\cmidrule{2-6}
			& w/ batch, Top-$N_b$& 87.1&  28& 93.6&    20\\
			    & w/ batch, DPP& \textbf{88.3}&   \textbf{25}& \textbf{94.2}&  \textbf{17}\\
			\bottomrule
		\end{tabular}
	\end{adjustbox}
	\label{tab:abl-batch}
\end{table}

\begin{table*}[hbt!]	
	\caption{Examples of the original and their adversarial sequences from Yelp and EC50 against BERT-base models.}
\label{tab:qualitative}
\centering
\begin{adjustbox}{max width=2\columnwidth}
		\begin{tabular}{llccc}
			\toprule
			%\multicolumn{2}{l}{Sentence-Level Text Classification (Movie Review)}&Label\\
			%\midrule

			%Orig  & suffers from a decided lack of creative storytelling.& Negative\\
			%\cmidrule(lr){2-2}
			%\cdashlinelr{2-2}
			%BBA  & \emph{\textcolor{red}{undergo}} from a decided \emph{\textcolor{red}{dearth}} of creative storytelling.& Positive\\
			%\cmidrule(lr){2-2}
			%\cdashlinelr{2-2}
			%TF    & -& Fail\\
			%\toprule
			%\multicolumn{2}{c}{Yelp}\\
			%\cmidrule(lr){1-3}
			%\midrule
			\multicolumn{2}{l}{Document-Level Text Classification (Yelp)}&Label\\
			\midrule
			Orig  & Food is fantastic and exceptionally clean! My only complaint is I went there with my 2 small children and they were showing a very&\multirow{2}{*}{Positive}\\ & inappropriate R rated movie! \\
			%\cmidrule(lr){2-2}
			\cdashlinelr{2-2}
			BBA  & Food is \emph{\red{gorgeous}} and exceptionally \emph{\red{unpolluted}}! My only complaint is I went there with my   2 small children and they were showing a very& \multirow{2}{*}{Negative}\\ & inappropriate R rated movie! \\
			%\cmidrule(lr){2-2}
			\cdashlinelr{2-2}
			TF    & Food is fantastic and \emph{\red{awfully}} clean! My only \emph{\red{grievances}} is I \emph{\red{turned}} there with my  2 small children and they were showing a very& \multirow{2}{*}{Negative}\\ & inappropriate R rated \emph{\red{footage}}! \\
			%\midrule
			%\multicolumn{2}{l}{Textual Entailment (QNLI)}&Label\\
		%\midrule
		%& Premise : What was the main idea of James Hutton's paper?\\
		%\cmidrule{2-2}
	
		%Orig & Hypothesis : In his \emph{\red{paper}}, he explained his theory that the Earth must be much older than had previously been supposed in order to& \multirow{3}{*}{Entailment}\\ & allow  enough time for mountains to be eroded and for sediments to form new rocks at the bottom of the sea, which in turn were\\ & raised up to become dry land.\\
			%\cdashlinelr{2-2}
			%BBA & Hypothesis : In his \emph{\red{journals}}, he explained his theory that the Earth must be much older than had previously been supposed in order to& \multirow{3}{*}{Not Entailment}\\ & allow enough time for mountains to be eroded and for sediments to form new rocks at the bottom of the sea, which in turn were\\ & raised up to become dry land.\\
			%\cdashlinelr{2-2}
			%TF &Hypothesis : In his paper, he explained his theory that the Earth must be much older than had previously been supposed in \emph{\red{writs}} to & \multirow{3}{*}{Not Entailment}\\ & \emph{\red{activation}} enough \emph{\red{timing}} for mountains to be eroded and for sediments to form new rocks at the bottom of the sea, which in turn were \\ &raised up to become dry land.\\
%	\bottomrule
%	\end{tabular}
%\end{adjustbox}
%\begin{adjustbox}{max width=2\columnwidth}
%	\begin{tabular}{ccccc}
%		\toprule
		\midrule
		\multicolumn{2}{l}{Protein Classification (EC50 level 0)}& Label\\
		\midrule
Orig&\texttt{MATPWRRALLMILASQVVTLVKCLEDDDVPEEWLLLHVVQGQIGAGNYSYLRLNHEGKIILRMQSLRGDADLYVSDSTPHPSFDDYELQSVT}&\multirow{3}{*}{Non-Enzyme}\\ &\texttt{CGQDVVSIPAHFQRPVGIGIYGHPSHHESDFEMRVYYDRTVDQYPFGEAAYFTDPTGASQQQAYAPEEAAQEEESVLWTILISILKLVLEILF} \\
			\cdashlinelr{2-2}
			BBA &\texttt{MATPWRRALLM\red{R}LASQVVTLVKCLEDDDVPEEWLLLHVVQGQIGAGNYSYLRLNHEGKIILRMQSLRGDADLYVSDSTPHPSFDDYELQSVT}& \multirow{3}{*}{Enzyme}\\ &\texttt{CGQDVVSIPAHFQRPVGIGIYGHPSHHESDFEMRVYYD\red{W}TVD\red{W}YPFGEAAYFTDPTGASQQQAYAPEEAAQEEESVLWTILISILKLVLEILF}\\
			\cdashlinelr{2-2}
			TF & \texttt{MATPWRRALLMILASQVVTLVKCLEDDDVPEEWLLLHVVQGQIGAGNYSYLRLNHEGKIILRMQSLRGDADLYVSDSTPHPSFDDYELQSVT}&\multirow{3}{*}{Enzyme}\\ &\texttt{CGQDVVSIPAHFQRPVGIGIYGHPSHHESDFEMRVYYDRTVDQYPFGE\red{W}AYF\red{C}\red{C}\red{G}\red{W}GASQQQAYAPEE\red{W}\red{W}\red{W}\red{F}EESVL\red{D}TILIS\red{G}LKLVLEILF} \\
\bottomrule
\end{tabular}
\end{adjustbox}
\end{table*}
\subsubsection{The Effect of Post-Optimization Process}
We analyze the trend of change in modification rate during the post-optimization process. %Since post-optimization process conducted after crafting adversarial sentence, this process is irrelvant to attack success rate. Instead, this process just reduce distortion between adversarial sentence and the original sentence using additional queries. Thus, post-optimization process has tradeoff between the number of queries and modification rate, and patience parameter in post-optimization process determine the degree of tradeoff. 
The post-optimization process reduces the distortion between the adversarial sequence and the original sequence using additional queries, which results in a trade-off between the distortion and the number of queries. 
\cref{fig:trav} shows the trajectory of the modification rate while traversing the query budget $N_\text{post}$ for post-optimization from $0$ to $200$. 
We find that the post-optimization process reduces the distortion and reaches the same distortion as PWWS using a fewer number of queries.
\begin{figure}[hbt!]
	\centering
	\begin{adjustbox}{max width=0.99\columnwidth}
		%\begin{subfigure}[t]{0.49\textwidth}
		\begin{tikzpicture}
		\begin{axis}[
		% Figure size
		width=4.5cm,
		height=4.2cm,
		% Plot style
		no marks,
		every axis plot/.append style={thick},
		% Grid
		grid=major,
		% Tick
		scaled ticks = false,
		ylabel near ticks,
		tick pos=left,
		tick label style={font=\small},
		xtick={0, 500,1000,1500,2000},
		xticklabels={0, 0.5k, 1k, 1.5k ,2k},
		ytick={0, 10,20,30},
		yticklabels={0, 10,20,30},
		% Label
		label style={font=\small},
		xlabel={Number of queries},
		ylabel={Modification rate (\%)},
		ylabel style={at={(-0.2,0.5)}},
		% Range
		xmin=0,
		xmax=2100,
		ymin=0,
		ymax=31,
		% Legend
		legend cell align={left},
		legend style={legend columns=1, at={(1.8, 0.62), mark size=1pt, font=\tiny}, font=\footnotesize},
		]
		\addplot[red, only marks, mark size=0.7pt] table [x=Qrs, y=modif, col sep=comma]{CSV_final/mp_trav_ours_wordnet.csv};
		\addlegendentry{BBA}
		\addplot[blue, only marks, mark size=0.7pt] table [x=PWWSQrs, y=PWWSmodif, col sep=comma]{CSV_final/mp_trav_baselines.csv};
		\addlegendentry{PWWS}
		\end{axis}
		\end{tikzpicture}
\end{adjustbox} 

\caption{Modification rate versus the number of queries plot of adversarial texts generated by traversing $N_\text{post}$ from $0$ to $200$ on the IMDB dataset against a BERT-base model. We use WordNet substitution method for the attack.}
	\label{fig:trav}
\end{figure}

\subsubsection{The Actual Runtime Analysis}

\begin{figure}[hbt!]
	\centering
	\begin{subfigure}[t]{0.65\columnwidth}
				\begin{tikzpicture}
			\begin{axis}[
				width=3.7cm, height=3.5cm, grid=major, no marks, scaled ticks = false, tick pos = left,  
			tick label style={font=\tiny}, 
			ytick={0,20000,40000,60000,80000,100000},
			yticklabels={0,20k,40k,60k,80k,100k},
			xtick={0,2000,4000,6000,8000,10000},
			xticklabels={0, 2k, 4k, 6k, 8k, 10k},
			label style={font=\tiny}, xlabel={Number of queries}, ylabel={Actual runtime (sec)}, ylabel style={at={(0.25,0.5)}}, xlabel style={at={(0.5,0.15)}}, xmin=0, xmax=10000, ymin=0, ymax=105000, legend style={legend columns=1, font=\tiny}, legend cell align={left}, legend style={at={(-1.04,0.53)},anchor=center}] %, legend pos=outer north east
			\addplot+[black] table [x=qrs, y=runtime, col sep=comma]{EXP2/bayesattack-hownet_categorical_inf_anal2_v3_straight_dpp_posterior_1000_pso_0-123.csv};
			\addlegendentry{w/o both}
			\addplot+[blue] table [x=qrs, y=runtime, col sep=comma]{EXP2/bayesattack-hownet_categorical_wide_anal2_v3_sod_straight_dpp_posterior_1000_pso_0-123.csv};
			\addlegendentry{w/ HS, $m\!=\!40$}
			\addplot+[red] table [x=qrs, y=runtime, col sep=comma]{EXP2/bayesattack-hownet_categorical_narrow_anal2_v3_sod_straight_dpp_posterior_1000_pso_0-123.csv};
			\addlegendentry{w/ HS, $m\!=\!5$}
			\end{axis}
			\end{tikzpicture}
			{\captionsetup{justification=raggedleft,singlelinecheck=false}
	\caption{$123$rd text~~~}}
				\label{fig:runtimea}
			\end{subfigure}
	\begin{subfigure}[t]{0.3\columnwidth}
			\begin{tikzpicture}
			\begin{axis}[width=3.7cm, height=3.5cm, grid=major, no marks, scaled ticks = false, tick pos = left, 
			tick label style={font=\tiny}, 
			ytick={0,20000},
			yticklabels={0,20k},
			xtick={0,2000,4000,6000,8000,10000},
			xticklabels={0, 2k, 4k, 6k, 8k, 10k},
			label style={font=\tiny}, xlabel={Number of queries}, xlabel style={at={(0.5,0.15)}}, xmin=0, xmax=10000, ymin=0, ymax=21000, legend style={legend columns=1, font=\tiny}, legend cell align={left}, legend style={at={(0.5,1.5)},anchor=center}] %, legend pos=outer north east
			\addplot+[black] table [x=qrs, y=runtime, col sep=comma]{EXP2/bayesattack-hownet_categorical_inf_anal2_v3_straight_dpp_posterior_1000_pso_0-348.csv};
			%\addlegendentry{w/o both}
			%\addplot+[red] table [x=qrs, y=runtime, col sep=comma]{EXP2/bayesattack-hownet_categorical_inf_anal2_v3_sod_straight_dpp_posterior_1000_pso_0-348.csv};
			%\addlegendentry{w/ HS}
			\addplot+[blue] table [x=qrs, y=runtime, col sep=comma]{EXP2/bayesattack-hownet_categorical_wide_anal2_v3_sod_straight_dpp_posterior_1000_pso_0-348.csv};
			%\addlegendentry{w/ HS, $B\!=\!40$}
			\addplot+[red] table [x=qrs, y=runtime, col sep=comma]{EXP2/bayesattack-hownet_categorical_narrow_anal2_v3_sod_straight_dpp_posterior_1000_pso_0-348.csv};
			%\addlegendentry{w/ HS, $B\!=\!5$}
			\end{axis}
			\end{tikzpicture}
	{\captionsetup{justification=raggedleft,singlelinecheck=false}
	\caption{$348$th text}}
	\label{fig:runtimeb}
\end{subfigure}
	\caption{The cumulative runtime versus the number of queries plot. HS in the legend denotes history subsampling, and $m=k$ in the legend denotes block decomposition with the block size $k$.}
	\label{fig:runtime}
\end{figure}
To study the effectiveness of block decomposition and history subsampling techinques on runtime, we choose two texts ($123$rd text of length $641$ and $348$th text of length $40$) from the texts that BBA iterates more than once until attack success when attacking the Yelp dataset against BERT-base model.
\Cref{fig:runtime} shows that block decomposition and history subsampling significantly reduces the actual runtime as the number of queries increases. 
Note that the comparison between the black and the blue curve of 348th text shows only the effect of history subsampling since the block size is equal to the text length (single block). In practice, attacking long documents against the robust model may require a large number of queries and our techniques can effectively reduce the actual runtime in that situation.

%\Cref{fig:runtime2} shows the cumulative distribution of the actual runtime required for attack methods when we use $20$\% of TextFooler's query budget as the query budget. \Cref{fig:runtime2} shows that BBA achieves $70$\% ASR which outperforms $40$\% ASR of TextFooler under the same query budget. %Also, except for the first $20$ seconds, BBA achieves higher ASR than TextFooler for the same runtime. 
\Cref{fig:runtime2} shows the cumulative distribution of the actual runtime required for attack methods. The result shows that BBA consistently finds successful texts faster than PWWS against the XLNet-large model on the Yelp dataset. Note that one could further accelerate the kernel computations of Bayesian optimization using a better computation resource such as a multi-GPU cluster. 
%The main focus of the paper was query efficiency, MR, and ASR. However, we think further enhancing actual runtime efficiency is an important future research.
\begin{figure}
	\centering
	\begin{subfigure}[b]{\columnwidth}
		\centering
		\begin{adjustbox}{max width=\columnwidth}
			\begin{tikzpicture}
				\begin{axis}[at={(0cm,-0.5cm)},width=4.5cm, height=4.2cm, grid=major, no marks, scaled ticks = false, tick pos = left,  
				tick label style={font=\small}, 
				ytick={0,0.2,0.4,0.6,0.8,1.0},
				yticklabels={0,20,40,60,80,100},
				xtick={0,100,200,400},
				label style={font=\small}, xlabel={Actual runtime (sec)}, ylabel={Success rate}, ylabel style={at={(0.1,0.5)}}, xlabel style={at={(0.5,0)}}, xmin=0, xmax=250, ymin=0, ymax=1.05, legend style={legend columns=1, font=\footnotesize}, legend cell align={left}, legend style={at={(1.7,0.5)},anchor=center}] %, legend pos=outer north east
				\addplot+[red] table [x=time, y=asr, col sep=comma]{TIME/xlnet-large-cased-yelp-ours-pwws.csv};
			\addlegendentry{BBA}
			\addplot+[blue] table [x=time, y=asr, col sep=comma]{TIME/xlnet-large-cased-yelp-pwws_0_product.csv};
			\addlegendentry{PWWS}
		\end{axis}
			\end{tikzpicture}
	\end{adjustbox}
	\end{subfigure}
	\caption{The cumulative distribution of the actual runtime required for the attack methods against the XLNet-large model on the Yelp dataset. Refer to \cref{tab:large} in \cref{app:add} for the detailed attack results.}
	\label{fig:runtime2}
\end{figure}
%In the case of the $123$rd text, we try $15000$ queries for $3$ days without BD and HS, but we cannot succeed in attacking. 
%However, we successfully attack the $123$rd text within 1 hour using BD and HS. 

%Since BO requires the computation of acquisition values for each step, the runtime per query of BO is longer than the runtime per query of the genetic or greedy algorithm.

%Even though BBA achieves a better attack success rate and MR using fewer queries compared to the baselines, the actual runtime of BBA is slower than the greedy methods due to overhead in the acquisition computation as shown in \Cref{fig:runtime}. That said, BBA can attack each text within 1.5 minutes on average, which is a reasonable speed in practice.

%Even though BBA achieves a better attack success rate and MR using fewer queries compared to the baselines, 
\subsection{Qualitative Results}
Attack examples of Yelp and EC50 datasets in \cref{tab:qualitative} show that our method successfully generates semantically consistent adversarial texts while baseline methods generate adversarial sequences with high modification rates. Please refer to \cref{app:qual} for more qualitative results.

\section{Conclusion}  
We propose a query-efficient and scalable black-box attack method on discrete sequential data using Bayesian optimization. 
In contrast to greedy-based state-of-the-art methods, our method can dynamically compute important positions using an ARD categorical kernel during Bayesian optimization. %, and consider the local relationship between elements in a sequence.
%which can alleviate the suboptimality of the greedy-based methods.  
Furthermore, we propose block decomposition and history subsampling techniques to scale our method to long sequences and large queries. 
Lastly, we develop a post-optimization algorithm that minimizes the perturbation size.
Our extensive experiments on various victim models and datasets from different domains show the state-of-the-art attack performance compared to the baseline methods. Our method achieves a higher attack success rate with a significantly lower modification rate and the number of queries throughout all our experiments. 

\section*{Broader Ethical Impact}
Our research focuses on the important problem of adversarial vulnerabilities of classification models on discrete sequential data. 
Even though there is the possibility of a malicious adversary misusing BBA to attack public text classification APIs, we believe our research can be a basis for the improvement in defenses against adversarial attacks on discrete sequential data.

\section*{Acknowledgements}
This work was supported by Samsung Research Funding \& Incubation Center of Samsung Electronics under Project Number SRFC-IT2101-01, 
Institute of Information \& communications Technology Planning \& Evaluation (IITP) grant funded by the Korea government (MSIT) (No. 2020-0-00882, (SW STAR LAB) Development of deployable learning intelligence via self-sustainable and trustworthy machine learning), 
and Basic Science Research Program through the National Research Foundation of Korea (NRF) (2020R1A2B5B03095585).
This material is based upon work supported by the Air Force Office of Scientific Research under award number FA2386-20-1-4043.
Hyun Oh Song is the corresponding author.

\bibliography{main}
\bibliographystyle{icml2022}

%%%%%%%%%%%%%%%%%%%%%%%%%%%%%%%%%%%%%%%%%%%%%%%%%%%%%%%%%%%%%%%%%%%%%%%%%%%%%%%
%%%%%%%%%%%%%%%%%%%%%%%%%%%%%%%%%%%%%%%%%%%%%%%%%%%%%%%%%%%%%%%%%%%%%%%%%%%%%%%
% APPENDIX
%%%%%%%%%%%%%%%%%%%%%%%%%%%%%%%%%%%%%%%%%%%%%%%%%%%%%%%%%%%%%%%%%%%%%%%%%%%%%%%
%%%%%%%%%%%%%%%%%%%%%%%%%%%%%%%%%%%%%%%%%%%%%%%%%%%%%%%%%%%%%%%%%%%%%%%%%%%%%%%
\newpage
\appendix
\onecolumn

\section{Algorithms}
\label{app:mainalg}
The overall algorithm of BBA is shown in \cref{alg:main}. For the ease of notation, we assume that $l$ is divisible by $m$. 

\begin{table}[hbt!]
	\small
	\centering
\begin{tabular}{lll}
	\toprule
	\multicolumn{2}{l}{Notations used in \cref{alg:main}}\\
	\midrule
$\mathcal{D}_k$&The evaluation history corresponding to the block $M_k$.\\	
$\mathcal{D}_k^{\text{sub}}$&The evaluation dataset subsampled from the evaluation history.\\
$E_k$&Exploration budget when optimizing the block $M_k$.\\
$N_k$&Query budget when optimizing the block $M_k$.\\
$N_{\text{post}}$&Query budget for the post-optimization process.\\
$N_{b}$&The batch size.\\
$\mathcal{S}(s',M_k)$ &The attack space when optimizing the block $M_k$ with an initial sequence $s'$.\\
& Formally, $\mathcal{S}(s',M_k) =  \{s''\in\prod_{i=0}^{l-1}\mathcal{C}(w_i')\mid w_i'' = w_i' \text{ for } i\notin M_k\}$. \\
$s_{\setminus M_k}$ & The subsequence of $s$ corresponding to the index set $[l]\setminus M_k$. Formally, $s_{\setminus M_k} = (w_i)_{i\in[l]\setminus M_k}$.\\
\bottomrule
\end{tabular}
\end{table}
\begin{algorithm*}[h]
    \caption{The overall algorithm of BBA.}
    \label{alg:main}
\begin{algorithmic}[1]
\STATE {\bfseries Input:} The original sequence $s$, its label $y$, blocks $\{M_k\}$, a target classifier $f_\theta$ and the attack criterion $\mathcal{L}$.
\STATE Initialize the current sequence $s_\text{cur} \leftarrow s$. 
\STATE Initialize the evaluation history $\mathcal{D}_k \leftarrow \emptyset$ for all $k=0,1,\ldots,l/m-1$.
\STATE Initialize the importance score $\alpha_k \leftarrow | \mathcal{L}(f_\theta(s),y) - \mathcal{L}(f_\theta(s_{\setminus M_k}),y)|$ for all $k=0,1,\ldots,l/m-1$.
\FOR{$\text{ITER}=0$ {\bfseries to} $R-1$}
	\STATE Sort the block indices $[0, \ldots , l/m-1]$ to $[\gamma_0, \ldots, \gamma_{l/m-1}]$ in order of decreasing importance score.
	\FOR{$k$ {\bfseries in} $[\gamma_0, \ldots, \gamma_{l/m-1}]$}
	\STATE Initialize the evaluation dataset $\mathcal{D}_k^\text{sub} \leftarrow \mathrm{SoD}(\mathcal{D}_k, N_k)$ (See \cref{alg:sod})
		\STATE Evaluate $E_k$ sequences randomly sampled from the attack space $\mathcal{S}(s_\text{cur},M_k)$ and append them to $\mathcal{D}_k^\text{sub}$ and $\mathcal{D}_k$.

		\STATE Update the remaining budget $N_r \leftarrow N_k - E_k$.
		\WHILE{$N_r>0$}
			\STATE Update $s_\text{cur}$ to the best sequence evaluated so far.
			\IF{$\mathcal{L}(f_\theta(s_\text{cur}), y)\ge 0$}
			\STATE {\bfseries Return} $\mathrm{PostOpt}(s, s_\text{cur},\mathrm{SoD}(\mathcal{D}_k, N_k),N_\text{post},N_b)$. (See \cref{alg:postopt})
		%\ELSIF{$\mathcal{L}(f_\theta(s_\text{best}), y) > \mathcal{L}(f_\theta(s_\text{cur}), y) $}
		%	\STATE Update current sequence $s_\text{cur} \leftarrow s_\text{best}$.
		\ENDIF
			\STATE Fit the GP parameters on $\mathcal{D}_k^\text{sub}$. (See \cref{app:GP})
			\STATE Select the sequence set $\mathcal{T}$ of top-$T$ acquisition value in $\mathcal{B}_H(s_\text{cur},1) \cap \mathcal{S}(s_\text{cur},M_k)$.
			\STATE Greedily select a batch $B$ of size $\min(N_b,N_r)$ from $\mathcal{T}$ that maximizes its DPP prior. (See \cref{app:acqmax})
			\STATE Evaluate the batch $B$ and append them to $\mathcal{D}_{k}^\text{sub}$ and $\mathcal{D}_{k}$. 
			%\STATE $s_\text{best} \leftarrow \argmax_{s'\in B} \mathcal{L}(f_\theta(s'), y)$.
			\STATE Update the remaining budget $N_r \leftarrow N_r - \min(N_b,N_r)$.	
		\ENDWHILE
	\ENDFOR
	\STATE Fit the GP parameters on $\bigcup_k\text{SoD}(\mathcal{D}_k, N_k)$.
	\STATE Update the importance score $\alpha_k \leftarrow \sum_{i\in M_k} 1/\beta_i$.
\ENDFOR
\STATE {\bfseries Return} The best sequence $s_\text{best}$ in $\bigcup_k \mathcal{D}_k$. 
\end{algorithmic}
\end{algorithm*}
\normalsize
\newpage
\twocolumn
\section{GP Surrogate Model Fitting}
\label{app:GP}

We use a GP surrogate model consisting of a constant mean function $\mu(\cdot;\eta)$ with mean $\eta$, an ARD categorical kernel $k^\text{cate}(\cdot,\cdot;\{\beta_i\},\sigma_f^2)$ with length-scale parameters $\{\beta_i\}$ and signal variance $\sigma_f^2$, and a homoskedastic noise with variance $\sigma_n^2$. Concretely, the surrogate model $g$ can be written by 
$$g(X)  \sim \mathcal{N}(\eta,K^\text{cate}(X,X;\{\beta_i\},\sigma_f^2)+\sigma_n^2 I).$$ 
For a given dataset $\mathcal{D}^\text{sub}$, we train the parameters $\eta$, $\{\beta_i\}$, $\sigma_f^2$, and $\sigma_n^2$ by maximizing the posterior probability distribution $p(\eta,\{\beta_i\},\sigma_f^2,\sigma_n^2\mid\mathcal{D}^\text{sub})$. Since 
\small
\begin{align*}
	&\overbrace{p(\eta,\{\beta_i\},\sigma_f^2,\sigma_n^2\mid\mathcal{D}^\text{sub})}^\text{Posterior probability}\\
	&~~~~~~~~~~~\propto p(\eta)\left(\prod_i p(\beta_i)\right) p(\sigma_f^2) p(\sigma_n^2) \underbrace{p(\mathcal{D}^\text{sub}|\eta,\{\beta_i\},\sigma_f^2,\sigma_n^2)}_{\text{Marginal likelihood}},
\end{align*}
\normalsize 
we maximize the sum of the log marginal likelihood of the dataset $\mathcal{D}^\text{sub}$ and the log prior probabilities of parameters $\eta$, $\{\beta_i\}$, $\sigma_f^2$, and $\sigma_n^2$. Concretely, we estimate $\eta$, $\{\beta_i\}$, $\sigma_f^2$, and $\sigma_n^2$ by solving
\small
\begin{align}
	&\maximize_{\eta,\{\beta_i\},\sigma_f^2,\sigma_n^2} ~ \log\left(p\left(\mathcal{D}^\text{sub}\mid\eta,\{\beta_i\},\sigma_f^2,\sigma_n^2\right)\right) + \log\left(p(\eta)\right) \nonumber\\&~~~~~~~+ \sum_i\log\left(p\left(
	\beta_i\right)\right) + \log\left( p\left(\sigma_f^2\right)\right) + \log\left(p\left(\sigma_n^2\right)\right).
	\label{eq:post}
\end{align}
\normalsize

For a more detailed explanation, we first introduce the prior distributions used in our experiments and explain the optimization procedure of  \cref{eq:post}.
\normalsize

\subsection{Prior on Parameters}
Here, we provide details about the prior distributions of GP parameters. 
We assume that the mean $\eta$ and the signal variance $\sigma_f^2$ have uniform priors on their domains. We assign an inverse gamma prior to the length-scale parameter $\beta_i$: $\beta_i^{-1}\sim\text{Gamma}(3.0,6.0)$, where $\text{Gamma}(a,b)$ is a gamma distribution with a concentration parameter $a$ and a rate parameter $b$. To induce the noise variance $\sigma_n^2$ to be close to zero, we use a gamma distribution with low concentration and high rate as a prior: $\sigma_n^2\sim\text{Gamma}(0.9,10.0)$.

\subsection{Optimization Method}
Since the log prior probabilities of GP parameters and log marginal likelihood are differentiable with respect to GP parameters, we use Adam, a gradient-based optimizer, to solve \cref{eq:post} \cite{Adam}. For each GP parameter fitting step, we run Adam with a learning rate of $0.1$ for $3$ iterations starting from the GP parameter derived at the previous fitting step (warm start).
%(For the initial GP parameter fitting step, we start from randomly initialized GP parameters, instead.) 
Then, we update the GP parameters with the values of the last iteration.

\section{Implementation Details}
\label{app:ID}

\subsection{Experimental Details}
\label{app:exp}

\subsubsection{Block Decomposition and History Subsampling}
\label{app:subsample}
As illustrated in \cref{fig:proc}, we subsample datasets for all blocks before starting the sequential block optimization step. When the optimization step ends, we fit the GP parameters to maximize the posterior probability distribution on $\bigcup_k\text{SoD}(\mathcal{D}_k, N_k)$ and reassign the importance score of the block $M_k$ to $\sum_{i\in M_k}1/\beta_i$.

For the experiments in the NLP domain, we fix the block size $m=40$. For the experiments in protein domain, we fix the block size $m=20$. To allocate a larger query budget to the blocks with larger search space, we fix $N_k = \sum_{i\in M_k} (|\mathcal{C}_i| - 1)$ in all experiments. 
\subsubsection{Acquisition Maximization}
\label{app:acqmax}
Here, we illustrate our batch update step in detail. We denote the evaluation dataset by $\mathcal{D}$ for notational simplicity. 
For each acquisition maximization step, we first find the sequence set $\mathcal{T}$ of top-$T$ acquisition values in the 1-Hamming distance ball $\mathcal{B}_H(s^*_\mathcal{D},1)$ of the best sequence $s^*_\mathcal{D}$ evaluated so far. Then, we initialize a batch $B = \{s_\mathcal{T}^*\}$, where $s_\mathcal{T}^*$ is the sequence with the highest acquisition value in $\mathcal{T}$. Finally, we repeatedly append the sequence $s^*$ that maximizes the marginal gain in the DPP prior to $B$ until the size of $B$ reaches $N_b$, which is represented as
\begin{align*}
	s^*&=\argmax_{s'\in\mathcal{T}\setminus B} \big(\det\left(\text{Var}\left(g(B\cup s'\mid\mathcal{D})\right)\right) 
	\\&~~~~~~~~~~~~~~~~~~~~- \det\left(\text{Var}\left(g(B\mid\mathcal{D})\right)\right)\big).
\end{align*}
To avoid redundant evaluations, we exclude sequences previously evaluated in the past acquisition maximization steps from the batch. We fix $N_b=4$ and $T=100$ for all experiments.
\subsubsection{Post-Optimization}
\label{app:postopt}

In the post-optimization process, we search an adversarial sequence in a reduced search space $\mathcal{B}_H(s,d_H(s,s_\text{adv})-1) \cap \mathcal{B}_H(s_\text{adv},r)$ to reduce the perturbation size.
We fix $r=2$ for all experiments. We choose the next sequence to evaluate by the following steps for efficiency. First, we randomly sample 300 sequences from $\mathcal{B}_H(s_\text{adv},2)\cap\mathcal{B}_H(s,d_H(s,s_\text{adv})-2)$ and find the sequence $s^*_\text{sampled}$ that has the largest acquisition value among them. Then, we choose the sequence $s^* \in \mathcal{B}_H(s^*_\text{sampled},1)$ that has the largest acquisition value in the $1$-Hamming distance ball of $s^*_\text{sampled}$. 

Since the cardinality of the $1$-Hamming distance ball of a sequence is linear to $l$, we evaluate the acquisition function $\mathcal{O}(l)$ times for each acquisition maximization step. Note that the brute-force acquisition maximization requires $\mathcal{O}(l^2)$ number of the acquisition function evaluations since the search space has the size quadratic to $l$.

%The post-optimization process uses additional queries to reduce the perturbation size of the adversarial sequence. The trade-off between the number of queries and the perturbation size can be controlled by the query budget $N_\text{post}$ of the post-optimization process. 
% We vary $N_\text{post}$. %to achieve a less perturbation size compared to the baseline methods. 
%Here, we emphasize that $N_\text{post}$ can be controlled without restarting BBA. 
\cref{tab:hyp1,tab:hyp2,tab:hyp3,tab:hyp4,tab:hyp5} show the values for the query budget of the post-optimization step $N_\text{post}$ used in our experiments.

\begin{table}[hbt!]
    \centering
	\caption{$N_\text{post}$ of BBA used in \cref{tab:main1} (AG's News, MR).}
\centering
\begin{adjustbox}{max width=\columnwidth}
\begin{tabular}{cccccc}
    \toprule
&$\mathcal{C}$&WordNet&Embedding&HowNet\\
\cmidrule(lr){3-3} \cmidrule(lr){4-4} \cmidrule(lr){5-5} 
Dataset&Model
& PWWS & TF & PSO\\ 
\midrule
AG&BERT-base&50&20&100\\
&LSTM&50&20&100\\
\midrule
MR&XLNet-base&50&20&100\\
&BERT-base&100&20&100\\
&LSTM&50&20&100\\
\bottomrule
\end{tabular}
\end{adjustbox}
	\label{tab:hyp1}
\end{table}
\begin{table}[hbt!]
    \centering
	\caption{$N_\text{post}$ of BBA used in \cref{tab:main2} (IMDB, Yelp).}
\centering
\begin{adjustbox}{max width=\columnwidth}
\begin{tabular}{cccccccc}
    \toprule
&$\mathcal{C}$&\multicolumn{2}{c}{WordNet}&\multicolumn{2}{c}{Embedding}&\multicolumn{2}{c}{HowNet}\\
\cmidrule(lr){3-4} \cmidrule(lr){5-6} \cmidrule(lr){7-8} 
Dataset&Model& PWWS&LSH & TF&LSH & PSO&LSH\\ 
\midrule
IMDB&BERT-base&100&50&20&50&50&100\\
\midrule
Yelp&BERT-base&200&50&20&100&50&100\\
\bottomrule
\end{tabular}
\end{adjustbox}
	\label{tab:hyp2}
\end{table}
\begin{table}[hbt!]
    \centering
	\caption{$N_\text{post}$ of BBA used in \cref{tab:large} (XLNet-large, BERT-large).}
\centering
\begin{adjustbox}{max width=\columnwidth}
\begin{tabular}{cccccc}
	\toprule
	&$\mathcal{C}$&WordNet&Embedding\\
	\cmidrule(lr){3-3} \cmidrule(lr){4-4}
	Dataset&Model
	& PWWS & TF \\ 
	\midrule
	IMDB&XLNet-large&100&50\\
	&BERT-large&150&50\\
	\midrule
	Yelp&XLNet-large&100&50\\
	&BERT-large&200&50\\
	\midrule
	MR&XLNet-large&100&20\\
	&BERT-large&200&20\\
	\bottomrule
\end{tabular}
\end{adjustbox}
	\label{tab:hyp4}
\end{table}
\begin{table}[hbt!]
    \centering
	\caption{$N_\text{post}$ of BBA used in \cref{tab:main3} (MNLI, QNLI).}
\centering
\begin{adjustbox}{max width=\columnwidth}
\begin{tabular}{cccccc}
	\toprule
	&$\mathcal{C}$&WordNet&Embedding&HowNet\\
	\cmidrule(lr){3-3} \cmidrule(lr){4-4} \cmidrule(lr){5-5} 
	Dataset&Model
	& PWWS & TF & PSO\\ 
	\midrule
	MNLI&BERT-base&150&50&150\\
	\midrule
	QNLI&BERT-base&150&50&150\\
	\bottomrule
\end{tabular}
\end{adjustbox}
	\label{tab:hyp3}
\end{table}
\begin{table}[hbt!]
    \centering
	\caption{$N_\text{post}$ of BBA used in \cref{tab:BAE} (BAE).}
\centering
\begin{adjustbox}{max width=\columnwidth}
\begin{tabular}{cccccc}
	\toprule
	Dataset&MR & AG & Yelp & IMDB\\
	\midrule
	$N_\text{post}$& 20&20&20&20\\
	\bottomrule
\end{tabular}
\end{adjustbox}
	\label{tab:hyp5}
\end{table}
For all experiments for the protein classfication task, we use $N_\text{post}=50$. 
% We do not use the post-optimization process in \cref{tab:abl-batch}. 

\subsection{Runtime Analysis}
\label{app:RA}
%BBA consists of two steps. For the first step, we find adversarial sequence using Bayesian optimization with several scalability techniques. Then, we additionally apply post-optimization process to the adversarial sentence to reduce the perturbation size. 

In this section, we explain the computational complexity of BBA. We first reiterate the computational complexity of the GP surrogate model fitting and the acquisition maximization in a generic Bayesian optimization. Then we analyze the complexity of each component of BBA. %the sequential block optimization process. Then, we analyze the complexity of the post-optimization process.
For simplicity, we assume that $|\mathcal{C}(w_i)|=C$ for all $i$ and $N_k = N$ for all $k$. 

As we noted in our main paper, GP surrogate model fitting requires the matrix inversion of the covariance matrix whose computational complexity is $\mathcal{O}(n_\text{data}^3)$, where $n_\text{data}$ is the cardinality of the dataset used in Bayesian optimization. When the inverse matrix of the covariance matrix is given, we can calculate the predictive mean and variance of a sequence by matrix-vector multiplications (see \cref{main:bo}), whose computational complexity is $\mathcal{O}(n_\text{data}^2)$. Thus, the computational complexity of the acquisition maximization step is $\mathcal{O}(n_\text{data}^2 N_\text{acq})$, where $N_\text{acq}$ is the number of acquisition function queries in the acquisition maximization step. 

\subsubsection{Sequential Block Optimization}
We note that $n_\text{data}$ is upper bounded by $O(N)$ when optimizing the block $M_k$ since we optimize $M_k$ with an initial dataset of size $N$ and maximum query budget $N$.
Hence, the GP surrogate model fitting step when optimizing $M_k$ requires $\mathcal{O}(N^3)$ computations. The number of the acquisition function queries is $\mathcal{O}(Cm)$ since we find the next sequence to evaluate by maximizing the acquisition function on the $1$-Hamming distance ball of the current best sequence. In summary, the sequential block optimization step requires $\mathcal{O}(N^3 + N^2Cm)$ computations for each query,  which is independent of $n$.

\subsubsection{History Subsampling}
We use the Subset of Data (SoD) method with Farthest Point Clustering (FPC) to select the evaluation dataset $\mathcal{D}_k^\text{sub}$ of size $N$ from the evaluation history $\mathcal{D}_k$. \cref{alg:sod} shows that SoD requires $\mathcal{O}(N |D_k^\text{sub}|)$ number of distance computations, resulting in $\mathcal{O}(N n/m)$ computational complexity. Hence, the history subsampling for all blocks has computational complexity of $\mathcal{O}(Nn)$, which is linear to $n$. 

\subsubsection{Importance Score Update}
When the optimization step ends, we fit the GP parameters using $\bigcup_k\text{SoD}(\mathcal{D}_k, N)$, whose cardinality is $\mathcal{O}(Nl/m)$. Hence, the importance score update requires $\mathcal{O}( Nn + N^3l^3/m^3 )$ computations, where each term corresponds to the history subsampling and the GP surrogate model fitting. Hence, the complexity of the importance score update is linear to $n$.

\subsubsection{Post-Optimization}
In the post-optimization process, we search an adversarial sequence in the reduced attack space to minimize the perturbation size. In the worst case, the number of queries required in the post-optimization process is $N_\text{post} d_H(s,s_\text{adv}^\text{(init)})$ which is upper bounded by $N_\text{post} l$. Here, $s_\text{adv}^\text{(init)}$ denotes the initial adversarial sequence of the post-optimization. Thus, $n_\text{data}$ is upper bounded by $O(Nm + N_\text{post}l)$. Also, $N_\text{acq}$ is independent of $n$ as explained in \cref{app:postopt}. In summary, the computational complexity of the post-optimization process for each query is independent of $n$.

\subsection{Target Datasets}
\label{app:datasets}
We evaluate BBA on 500 samples randomly selected from the test set throughout all the experiments, following the protocol in \citet{LSH}.

\subsubsection{Text Classification}
To show the wide applicability of BBA, we evaluate BBA on various text classification datasets, including both sentence and document-level datasets.

$\bullet$ \textbf{AG's News} \cite{Yelp_and_AG}: a sentence-level dataset for classifying a news-type sentence into 4 topics: World, Sports, Business, and Science \& Tech. \\
$\bullet$ \textbf{Movie Review} \cite{MR}: a sentence-level sentiment classification dataset which consists of positive and negative movie reviews from Rotten Tomatoes.\\
$\bullet$ \textbf{IMDB Polarity} \cite{IMDB}: a document-level dataset for binary sentiment classification, which consists of highly polar movie reviews from IMDB. \\
$\bullet$ \textbf{Yelp Polarity} \cite{Yelp_and_AG}: a document-level dataset for binary sentiment classification, which consists of highly polar restaurant reviews.

\subsubsection{Textual Entailment}
We also conduct experiments on textual entailment datasets. 

$\bullet$ \textbf{MNLI matched} \cite{MNLI}: a textual entailment dataset, which consists of sentence pairs from transcribed speech, popular fiction, and government reports. 
The task is to judge the relationship between a pair of sentences, premise, and hypothesis. The test and training sets are derived from the same sources.\\
$\bullet$ \textbf{QNLI} \cite{QNLI}: a textual entailment dataset, which consists of question-sentence pairs converted from SQuAD v1.1 dataset \cite{SQuAD}. The task is to determine whether the sentence contains the answer to the given question or not.

Note that we only perturb hypotheses in MNLI and sentences in QNLI.

\subsubsection{Protein Classification}
Furthermore, we consider a protein classification dataset to show that BBA is also applicable to tasks other than NLP tasks. 
We note that a protein is a sequence of amino acids, each of which is a discrete categorical variable.

$\bullet$ \textbf{EC50} \cite{EC50}: an enzyme multi-label classification dataset generated by clustering amino acid sequences in Swiss-Prot database \cite{swissprot}. 
Each amino acid sequence in EC50 has three hierarchical labels, which correspond to enzyme versus non-enzyme (level 0, 2 classes), main enzyme class (level 1, 6 classes), and enzyme subclass (level 2, 65 classes), respectively. 

Each sequence in EC50 is coded with 28 symbols consisting of 25 amino acids from \citet{EC50} and 3 auxiliary tokens. The symbols are summarized in \cref{tab:symbols}.
\begin{table}
	\centering
	\begin{tabular}{cl}
		\toprule
Symbol & Amino acid\\
\midrule
A	&Alanine\\
R	&Arginine\\
N	&Asparagine\\
D	&Aspartic acid\\
C	&Cysteine\\
Q	&Glutamine\\
E	&Glutamic acid\\
G	&Glycine\\
H	&Histidine\\
I	&Isoleucine\\
L	&Leucine\\
K	&Lysine\\
M	&Methionine\\
F	&Phenylalanine\\
P	&Proline\\
O	&Pyrrolysine\\
S	&Serine\\
U	&Selenocysteine\\
T	&Threonine\\
W	&Tryptophan\\
Y	&Tyrosine\\
V	&Valine\\
B	&Aspartic acid or Asparagine\\
Z	&Glutamic acid or Glutamine\\
X	&Any amino acid\\
\_\_bos\_\_ & Beginning of a sentence (BOS) token\\
\_\_mask\_\_ & Mask token\\
\_\_pad\_\_ & Pad token \\
\bottomrule
	\end{tabular}
	\caption{The description of the 28 symbols used in EC50 dataset.}
	\label{tab:symbols}
\end{table}

%The categories of token 
%A sequence in 
%An amino acid of a protein in EC50 dataset has one of 25 categories.

\subsection{Victim Models}
\label{app:models}
For the sentence-level text classification datasets (AG's News, Movie Review), we consider multiple types of victim models to attack, including bi-directional word LSTM, fine-tuned BERT-base, BERT-large, XLNet-base, and XLNet-large \cite{LSTM, BERT, XLNet}. For the document-level text classification datasets, we focus on fine-tuned BERT-base,  BERT-large, and XLNet-large as the victim model. For the textual entailment datasets, we focus on fine-tuned BERT-base as the victim model. We fine-tune BERT-large and XLNet-large models following the training protocol in TextAttack API (refer to our Github repository). For other models, we utilize publicly released models from Hugging Face and TextAttack Framework \cite{huggingface, TextAttack}, following the practice in \citet{LSH}. The average text length and the original classification accuracy of the victim models we consider for each dataset are summarized in \cref{datastat}. 

For the protein classification dataset, we train three ASGD Weight-Dropped LSTM (AWD-LSTM) models \citep{AWD-LSTM}, each corresponding to the classification task for one of the three levels, following the training protocol of \citet{EC50}. Then, we consider these AWD-LSTM models as victim models. We note that the average length of the test samples selected from EC50 is 411.3 and the original classification accuracy of the victim models for levels 0, 1, and 2 are 96.0\%, 93.2\%, and 92.4\%, respectively. 

\begin{table}[ht]	
	\caption{The average text length and the classification accuracy of the victim models for each NLP dataset.}
	\label{datastat}
	\centering
	\begin{adjustbox}{max width=\columnwidth}
		\begin{tabular}{cccc}
			\toprule
			Dataset&Model&Avg. Len.&Acc. (\%)\\
			\midrule
			AG&BERT-base&39.0&93.8\\
			&LSTM&&90.4\\
			\midrule
			MR&XLNet-large&18.2&88.8\\
			&BERT-large&&84.8\\
			&XLNet-base&&87.2\\
			&BERT-base&&83.4\\
			&LSTM&&78.6\\
			\midrule
			IMDB&XLNet-large&242.5&96.0\\
			&BERT-large&&93.8\\
			%&XLNet-base&&94.2\\
			&BERT-base&&91.4\\
			\midrule
			Yelp&XLNet-large&136.0&98.6\\
			&BERT-large&&98.6\\
			%&XLNet-base&&98.8\\
			&BERT-base&&97.8\\
			\midrule
			MNLI&BERT-base&29.7&87.4\\
			\midrule
			QNLI&BERT-base&37.6&92.2\\
			\bottomrule
			\end{tabular}
	\end{adjustbox}
\end{table}

\subsection{Details about Comparison in NLP}
\label{app:spaces}

\begin{table}[ht]
	\caption{Attack results of the original TextFooler (TF-orig) and TextFooler-fixed (TF-fixed) on the setence-classification datasets.}
	\centering
		\begin{adjustbox}{max width=\columnwidth}
	\begin{tabular}{cccccc}
		\toprule
		Model&Dataset&Method&ASR (\%)&MR (\%) & Qrs\\
		\midrule
		BERT-base&AG&   TF-orig& 84.2& 24.6&   342\\
		&    &   TF-fixed& 84.7& 24.9&   346\\
		\cmidrule{2-6}
		&MR&   TF-orig& 88.7& 20.0&   115\\
		&    &   TF-fixed& 89.2& 20.0&   115\\
		\midrule
	LSTM&AG&   TF-orig& 94.7& 17.4&   229\\
		&    &   TF-fixed& 94.9& 17.3&   228\\
		\cmidrule{2-6}
		&MR&   TF-orig& 98.2& 13.6&    72\\
		&    &   TF-fixed& 98.2& 13.6&    72\\
		\midrule
	XLNet&MR&   TF-orig& 95.2& 18.1&   101\\
		&    &   TF-fixed& 95.0& 18.0&   101\\
		\bottomrule
	\end{tabular}
	\end{adjustbox}
	\label{tab:fixtf}
	\end{table}
	\label{app:tf}
	\begin{table}[ht]
		\caption{Attack results of the original TextFooler (TF-orig) and TextFooler-fixed (TF-fixed) on the document-classification datasets against BERT-base models.}
		\centering
			\begin{adjustbox}{max width=\columnwidth}
		\begin{tabular}{cccccc}
			\toprule
	Model&Method&ASR (\%) & MR (\%) &Qrs \\
	\midrule
	IMDB&   TF-orig& 98.9& 8.5&   706\\
			&   TF-fixed& 99.1& 8.6&   712\\
	\midrule
			Yelp&   TF-orig& 92.8& 10.8&   460\\
			&   TF-fixed& 93.5& 11.1&   461\\
		\bottomrule
	\end{tabular}
	\end{adjustbox}
	\label{tab:fixtf2}
	\end{table}
Here, we explain how we achieve a fair comparison with the baseline methods, focusing on the word substitution methods and attack spaces. 

\begin{table*}[hbt!]
    \centering
\caption{Attack results against XLNet-large and BERT-large on IMDB, Yelp, and MR.}
\begin{subtable}[h!]{0.9\columnwidth}
	\label{tab:large}
	\caption{WordNet}
\centering
\begin{adjustbox}{max width=\columnwidth}
\begin{tabular}{cccccc}
    \toprule
    Dataset&Model&Method & ASR (\%)& MR (\%)& Qrs \\
	\midrule
	IMDB&XLNet-large& PWWS& 98.8& 5.7&    1697 							 \\ %%
		& 		    &  BBA& \textbf{99.8}& \textbf{5.4} &    \textbf{573} \\ %% 100
		\cmidrule(r){2-6}
		&BERT-large & PWWS& 98.1 & 4.9 &   1661 \\ %%
        & 		    &  BBA& \textbf{100.0}& \textbf{4.6} &    \textbf{619} \\ %% 150
        %&XLNet-base & PWWS& 98.3& 5.2 &    1672							 \\ %%
        %& 		    &  BBA& \textbf{99.8}& \textbf{4.7} &    \textbf{513} \\ %% 100         
	\midrule
	Yelp&XLNet-large& PWWS& 94.5& 10.8&    1107 \\ 
		& 		    &  BBA& \textbf{98.2}& \textbf{9.4} &    \textbf{485} \\ %% 100
		\cmidrule(r){2-6}
		&BERT-large & PWWS& 75.1 & \textbf{7.7} &   1206 \\ %%
        & 		    &  BBA& \textbf{94.7}& \textbf{7.7} &    \textbf{657} \\ %% 200
        %&XLNet-base & PWWS& 91.1& 9.0 &    1099 \\ %%
        %& 		    &  BBA& \textbf{97.4}& \textbf{8.4} &    \textbf{441} \\ %% 100         
	\midrule
	MR  &XLNet-large& PWWS& 82.7& 14.9&    145 \\ %%
		& 		    &  BBA& \textbf{87.6}& \textbf{14.8} &    \textbf{98} \\ %% 100
		\cmidrule(r){2-6}
		&BERT-large & PWWS& 80.0 & 14.6 &   146 \\ %%
        & 		    &  BBA& \textbf{85.1}& \textbf{14.2} &    \textbf{110} \\ %% 200
	\bottomrule 
\end{tabular}
\end{adjustbox}
\end{subtable}
~~~~
\begin{subtable}[h!]{0.883\columnwidth}
	\caption{Embedding}
	\centering
	\begin{adjustbox}{max width=\columnwidth}
	\begin{tabular}{cccccc}
		\toprule
    Dataset&Model&Method & ASR (\%)& MR (\%)& Qrs \\
	\midrule
	IMDB&XLNet-large&   TF& 99.0& 8.8&    789 							 \\ %% 
		& 		    &  BBA& \textbf{99.6}& \textbf{6.9} &    \textbf{638} \\ %% 50 
		\cmidrule(r){2-6}
		&BERT-large &   TF& 97.2 & 7.0 &   660 \\ %%
        & 		    &  BBA& \textbf{98.7}& \textbf{6.4} &    \textbf{559} \\ %% 50
        %&XLNet-base &   TF& 99.4& 7.1 &   627 							 \\ %%
        %& 		    &  BBA& \textbf{100.0}& \textbf{6.0} &    \textbf{534} \\ %% 50
	\midrule
	Yelp&XLNet-large&   TF& 95.9& 16.7&    635 \\ %%
		& 		    &  BBA& \textbf{99.4}& \textbf{12.6} &    \textbf{475} \\ %% 50
		\cmidrule(r){2-6}
		&BERT-large &   TF& 68.8 & 14.1 &   924 \\ %%
        & 		    &  BBA& \textbf{93.9}& \textbf{11.3} &    \textbf{548} \\ %% 50 
        %&XLNet-base &   TF& 89.7& 12.1 &    599 \\ %% 
        %& 		    &  BBA& \textbf{98.0}& \textbf{10.9} &    \textbf{449} \\ %% 50  
	\midrule
	MR  &XLNet-large&   TF& 94.6& 18.4&    103 \\ %%
		& 		    &  BBA& \textbf{96.9}& \textbf{16.2} &    \textbf{68} \\ %% 20
		\cmidrule(r){2-6}
		&BERT-large &   TF& 90.8 & 18.7 &   111 \\ %%
        & 		    &  BBA& \textbf{94.8}& \textbf{17.0} &    \textbf{71} \\ %% 20
	\bottomrule 
\end{tabular}
\end{adjustbox}
\end{subtable}
\end{table*}
\subsubsection{Comparison with PWWS and PSO}
PWWS utilizes the word substitution method based on WordNet, which is a lexical database for the English language containing synonym sets for English words. Note that the WordNet-Based synonym set for a word $w$ is built without considering the context of the word in the text. Thus, we can write the synonym set by $\mathcal{C}^\text{wordnet}(w)$. For a fair comparison, we compare BBA and PWWS with the same word synonym sets. Specifically, we apply BBA on the attack space defined by the product space of WordNet-based synonym sets, $\prod_{i=0}^{l-1}\mathcal{C}^\text{wordnet}(w_i)$.

PSO proposes a word-substitution model based on the semantic labels of words, also referred to as sememes, and applies particle swarm optimization on the product space of sememe sets corresponding to the original words. For a fair comparison, we apply BBA on the same attack space with PSO. 

\subsubsection{Comparison with TextFooler}
TextFooler proposes an embedding-based synonym set that considers both word and sentence similarities. Thus, we can express the synonym set of a word $w'$ as a function of the word and the sentence $s'$ containing the word, denoted by $\mathcal{C}^\text{Emb}(w', s')$. Note that the synonym set can change dynamically during the optimization.

We first suggest TextFooler-fixed, a variant of the TextFooler method whose synonym set is defined by $\mathcal{C}^{\text{EmbFix}}(w')\triangleq \mathcal{C}^\text{Emb}(w',s)$, where $s$ is the original input sentence. As defined above, the synonym set of TextFooler-fixed is fixed during the optimization. \cref{tab:fixtf,tab:fixtf2} shows that TextFooler-fixed achieves attack performance comparable to the original TextFooler method. For a fair comparison, we apply BBA on the product space $\prod_{i=0}^{l-1}\mathcal{C}^\text{EmbFix}(w_i)$ and compare BBA with the TextFooler-fixed method instead of the original TextFooler method to use the same synonym sets for the attack.

\subsubsection{Comparison with LSH}
We compare BBA with the LSH method using each of all three word substitution methods, WordNet, Embedding, and HowNet. For experiments using the Embedding-based word substitution method, we set the attack space to the same as the TextFooler-fixed method.

%\begin{table}[ht]
%	\caption{Attack results of the original TextFooler (TF-orig) and TextFooler-fixed (TF-fixed) on the %document-classification datasets.}
%	\centering
%		\begin{adjustbox}{max width=\columnwidth}
%	\begin{tabular}{cccccc}
%		\toprule
%Model&Method&ASR (\%) & MR (\%) &Qrs \\
%\midrule
%IMDB&  LSH-orig& 98.47& 5.00&   774\\
%    &  LSH-fixed& 98.47& 4.96&   770\\
%	\midrule
%Yelp&  LSH-orig& 94.27& 8.82&   552\\
%    &  LSH-fixed& 94.68& 8.94&   550\\
%	\bottomrule
%\end{tabular}
%\end{adjustbox}
%\label{tab:fixlsh}
%\end{table}

\section{Additional Results}
\label{app:AR}

\subsection{Additional Baselines and Target Models}
\label{app:add}
\Cref{tab:large} shows that BBA outperforms PWWS and TextFooler in Movie Review, Yelp, and IMDB datasets against BERT-large and XLNet-large models. 
Also, BBA achieve state-of-the-art performance in all three evaluation metrics on textual entailment tasks, as shown in \cref{tab:main3}. 
\Cref{tab:BAE} shows that BBA outperforms \citet{BAE} (BAE) in various datasets against BERT-base.
\begin{table}[ht]
    \centering
\caption{Attack results for BERT-base models on textual entailment datasets. $\mathcal{C}$ denotes the word substitution method used in attack. }
\begin{subtable}[ht]{\columnwidth}
	\label{tab:main3}
	\centering
\begin{adjustbox}{max width=\columnwidth}
\begin{tabular}{cccccc}
    \toprule
    $\mathcal{C}$&Dataset&Method & ASR (\%)& MR (\%)& Qrs \\
	\midrule
	WordNet&MNLI& PWWS& 84.4 & 6.1 &   102 \\
    &    & BBA& \textbf{85.1}& \textbf{5.9} &    \textbf{42} \\ %% 150
\cmidrule{2-6}
&QNLI& PWWS& 67.7 & 9.3 &   211 \\
&    & BBA& \textbf{73.3} & \textbf{8.8} &   \textbf{168} \\ %% 150
\midrule
Embedding&MNLI&   TF& 92.5 & 7.3 &    77\\
&    & BBA& \textbf{93.8}& \textbf{6.2}&    \textbf{49}\\ %% 50
\cmidrule{2-6}
&QNLI&   TF& 83.7& 11.1 &   172\\
&    & BBA& \textbf{87.0}& \textbf{9.4}&   \textbf{157}\\ %% 50
	\midrule 
HowNet&MNLI&  PSO& \textbf{85.6} & 6.2 &   951\\
&    & BBA& \textbf{85.6}& \textbf{5.6}&    \textbf{100} \\ %% 150
\cmidrule{2-6}
&QNLI&  PSO& 68.6 & 13.8 & 31642 \\
&    & BBA& \textbf{70.1} & \textbf{9.2}&  \textbf{2369} \\ %% 150
	\bottomrule
\end{tabular}
\end{adjustbox}
\end{subtable}
\end{table}
\begin{table}[ht]
	\centering
	\caption{Comparison with BAE against BERT-base model on various datasets.}
	\begin{adjustbox}{max width=\columnwidth}
	\begin{tabular}{cccccc}
	\toprule
	Dataset&Method&ASR (\%)& MR (\%)&Qrs \\%&Avg runtime (sec)\\
	\midrule
	MR&BAE&81.3&16.1&87\\%&\textbf{3.2}\\
	&BBA&\textbf{93.1}&\textbf{14.2}&\textbf{57}\\ %% 20
	\midrule
	AG&BAE&53.7& 23.1& 329\\%, 13.83\\
	&BBA&\textbf{73.6}& \textbf{21.6}& \textbf{246}\\ %% 20
	\midrule
	Yelp&BAE&88.1&11.5&402\\%&\textbf{36.7}\\
	&BBA&\textbf{97.8}&\textbf{10.1}&\textbf{266}\\ %% 20
	\midrule
	IMDB&BAE&95.2& 8.9& 592\\%, 70.29\\
	&BBA&\textbf{98.9}& \textbf{5.3}& \textbf{353}\\ %% 20
	\bottomrule
	\end{tabular}
	\end{adjustbox}
	\label{tab:BAE}
\end{table}

\subsection{The Effect of Varying Block Size w/, w/o Post-Optimization}
\begin{table}[ht]
	\center
	\small
	\caption{Attack results for a AWD-LSTM model on EC50 level 1 dataset while varying the block size $m$.}
	\begin{adjustbox}{max width=\columnwidth}
		\begin{tabular}{ccccccc}
			\toprule
			&&\multicolumn{2}{c}{w/o Post-optimization}&\multicolumn{2}{c}{w/ Post-optimization}\\
			\cmidrule(lr){3-4}\cmidrule(lr){5-6}
			 $m$ &ASR (\%) & MR (\%)	&Qrs  & MR (\%) & Qrs\\
			\midrule
			2	&94.6 	&3.5	&186	&  1.7 & 428\\
			5	&97.6 	&4.6	&137	&  1.8 & 315\\
			10  &99.6 	&6.2	&107	&  2.1 & 272\\
			20  &99.8 	&9.3	&105	& 2.3 & 293\\
			40  &99.6 	&15.0	&86	 & 2.5 & 326\\
			80  &100.0 	&25.1	&57  & 2.6 & 403\\
			\bottomrule
		\end{tabular}
	\end{adjustbox}
	\label{tab:mtrav}
\end{table}
%Here, we explain that BBA with too large block size $m$ rather shows worse performance, even though it requires more runtime per queries in average. To this end, 
%In this section we analyze the effect of block size $m$ in BBA. 
We investigate the effect of varying the block size $m$ on the evaluation metrics for the protein classification task, with and without the post-optimization step.
 % with and without post-optimization process of the same query budget $N_\text{post}$. 
%For a fixed number of queries, the smaller block size $m$ leads to a smaller number of queries for each block, which makes the optimization algorithm run faster in practice.
%However, if $m$ is small, the algorithm cannot capture the relationship between the elements, which might result in a suboptimal solution. 
\cref{tab:mtrav} shows that BBA with a larger block size achieves a higher attack success rate with fewer queries but higher modification rate. 
%We interpret that when the block size is small, the local relationship is not well considered, resulting in suboptimality. 
The attack results with the post-optimization process show that the post-optimization successfully reduces the modification rate.%, but requires a larger number of queries when the block size increases.
%\blue{The attack results with the post-optimization process show that BBA with block size $20$ achieves lower distortion using fewer number of queries compared to BBA of block size $40$ and $80$ while BBAs with larger block sizes require longer runtime per queries in average.} %This result exhibits that larger block sizes are not always more query efficient.

\subsection{Query Efficiency of BBA}
We provide more cumulative distribution plots against BERT-base models on Movie Review, IMDB, and MNLI datasets. \cref{fig:cdmr,fig:cdimdb,fig:cdmnli} show that BBA can find the adversarial sequences using fewer queries than the baseline methods also on the Movie Review, IMDB, and MNLI datasets.

\input{fig5.tex}
%\input{fig4.tex}
\input{fig3.tex}
\input{fig6.tex}

\subsection{Qualitative Results}
\label{app:qual}
%Attack examples of Movie Review, Yelp, IMDB, EC50, MNLI, and QNLI datasets in \cref{tab:qualitative1}, \cref{tab:qualitative2}, and \cref{tab:qual3} show that BBA successfully generates semantically consistent adversarial texts while baseline methods fail to attack or generate adversarial sequences with high modification rates. 
Attack examples of Movie Review, Yelp, IMDB, MNLI, and QNLI datasets in \cref{tab:qualitative1,tab:qual3} show that BBA successfully generates semantically consistent adversarial texts while baseline methods fail to attack or generate adversarial sequences with high modification rates. 

\begin{table*}[hbt!]
	\centering	
	\caption{Examples of the original and their adversarial sequences from MR, Yelp, and IMDB against BERT-base models.}
\label{tab:qualitative1}
\begin{adjustbox}{max width=2\columnwidth}
		\begin{tabular}{clccc}
			\toprule
			\multicolumn{2}{l}{Sentence-Level Text Classification (Movie Review)}&Label\\
			\midrule
	Orig  & suffers from a decided lack of creative storytelling.& Negative\\
			%\cmidrule(lr){2-2}
			\cdashlinelr{2-2}
			BBA  & \emph{\textcolor{red}{undergo}} from a decided \emph{\textcolor{red}{dearth}} of creative storytelling.& Positive\\
			%\cmidrule(lr){2-2}
			\cdashlinelr{2-2}
			TF    & -& Fail\\
			%\toprule
			%\multicolumn{2}{c}{Yelp}\\
			%\cmidrule(lr){1-3}
			\midrule
			\multicolumn{2}{l}{Document-Level Text Classification (Yelp)}&Label\\
			\midrule
			Orig  & I had never been here before, but I'm glad I tried it out. It was the best massage that I've ever had. &Positive\\
			\cdashlinelr{2-2}
			%\cmidrule(lr){2-2}
			BBA  & I had never been here before, but I'm glad I tried it out. It was the \emph{\red{allright}} \emph{\red{massaging}} that I've ever had.& Negative\\
			\cdashlinelr{2-2}
			%\cmidrule(lr){2-2}
			TF    & I had never been here before, but I'm \emph{\red{excited}} \emph{\red{me}} tried it out. \emph{\red{His}} was the \emph{\red{alright}} massage that I've ever \emph{\red{received}}.& Negative\\
			\midrule
			\multirow{2}{*}{Orig}  & Not sure if they are closed for business but none of my calls were returned when I left a few VM's. &\multirow{2}{*}{Negative}\\ &Will update this if they ever do call back.\\
			\cdashlinelr{2-2}
			%\cmidrule(lr){2-2}
			\multirow{2}{*}{BBA}  & Not sure if they are closed for business but none of my \emph{\red{ask}} were returned when I left a few VM's. & \multirow{2}{*}{Positive}\\ & Will \emph{\red{refreshing}} this if they ever do call back.\\
			\cdashlinelr{2-2}
			%\cmidrule(lr){2-2}
			\multirow{2}{*}{TF}    & Not sure if they are closed for \emph{\red{companies}} but none of my \emph{\red{appeals}} were returned when I \emph{\red{going}} a \emph{\red{short}} VM's. & \multirow{2}{*}{Positive}\\ & \emph{\red{Dedication}} update this if they ever do call back.\\
			%\midrule
			%\multicolumn{2}{l}{Document-Level Text Classification (Yelp)}&Label\\
			%\midrule
			%Orig  & Food is fantastic and exceptionally clean! &\multirow{2}{*}{Positive}\\ &  My only complaint is I went there with my 2 small children and they were showing a very inappropriate R rated movie! \\
			%\cmidrule(lr){2-2}
			%\cdashlinelr{2-2}
			%BBA  & Food is \emph{\red{gorgeous}} and exceptionally \emph{\red{unpolluted}}! & \multirow{2}{*}{Negative}\\ & My only complaint is I went there with my   2 small children and they were showing a very inappropriate R rated movie! \\
			%\cmidrule(lr){2-2}
			%\cdashlinelr{2-2}
			%TF    & Food is fantastic and \emph{\red{awfully}} clean! & \multirow{2}{*}{Negative}\\ & My only \emph{\red{grievances}} is I \emph{\red{turned}} there with my  2 small children and they were showing a very inappropriate R rated \emph{\red{footage}}! \\
			\midrule
			\multicolumn{2}{l}{Document-Level Text Classification (IMDB)}&Label\\
			\midrule
			\multirow{6}{*}{Orig} &  
			I rented this movie, because I noticed the cover in the video rental store. & \multirow{6}{*}{Negative}\\ &I saw Nolte, Connely, Madsen, 40's time setting, and thought ``hmm, can't be too bad.''\\ & Unfortunately, after watching it, my impression was ``not too good''. \\ & Its kind of a Chinatown ripoff, but the worst part is that other than Nolte, the other members of the squad didn't get enough screen time. \\ & But its a decent movie to see once I guess. \\ & And Melanie's role  was small enough that she wasn't given a chance to be a nuisance. \\
			\cdashlinelr{2-2}
			%\cmidrule(lr){2-2}
			\multirow{6}{*}{BBA}  & 
			I rented this movie, because I noticed the cover in the video rental store. & \multirow{6}{*}{Positive}\\ &I saw Nolte, Connely, Madsen, 40's time setting, and thought ``hmm, can't be too bad.''\\ & \emph{\red{Unluckily}}, after watching it, my impression was ``not too good''.\\ & Its kind of a Chinatown ripoff, but the worst part is that other than Nolte, the other members of the squad didn't get enough screen time. \\ & But its a \emph{\red{honest}} movie to see once I guess. \\ & And Melanie's role was small enough that she wasn't given a chance to be a nuisance.\\
			\cdashlinelr{2-2}
			%\cmidrule(lr){2-2}
			\multirow{6}{*}{PSO}  & I rented this \emph{\red{documentary}}, because I noticed the cover in the video rental store. & \multirow{6}{*}{Positive} \\ &I \emph{\red{snapped}} Nolte, Connely, Madsen, 40's time setting, and thought ``hmm, can't be too bad.''\\  & \emph{\red{Unluckily}}, after watching it, my \emph{\red{career}} was ``not too good''. \\ & Its kind of a Chinatown ripoff, but the \emph{\red{seediest}} \emph{\red{farewell}} is that other than Nolte, the other members of the \emph{\red{world}}  didn't get \emph{\red{substantial}} screen time. \\  & But its a \emph{\red{honest}} \emph{\red{ballad}} to see once I guess. \\ & And Melanie's role was \emph{\red{humble}} enough that she wasn't given a \emph{\red{opportunity}} to be a \emph{\red{headache}}.\\
			\bottomrule
			
\end{tabular}
\end{adjustbox}
\end{table*}

\begin{table*}[hbt!]
	\centering	
	\caption{Examples of the original and their adversarial sequences from MNLI and QNLI against BERT-base models.}
	\begin{adjustbox}{max width=2\columnwidth}
		\begin{tabular}{clccc}
			\toprule
			%\multicolumn{3}{l}{Dataset}&Label\\
			\multicolumn{2}{l}{Textual Entailment (MNLI)}&Label\\
			\midrule
			Premise &  that's really true a lot of it is um the color certain colors seem to be more acceptable. \\
			\cmidrule{2-4}
			Orig& It doesn't make a difference what color one wears. & Neutral\\
			\cdashlinelr{2-2}
			BBA& It doesn't make a \emph{\red{dispute}} what color one wears.& Entailment \\
			\cdashlinelr{2-2}
			TF& \emph{\red{His}} doesn't \emph{\red{do}} a difference what \emph{\red{colourful}} one \emph{\red{focuses}}. & Entailment\\
			\midrule
			Premise &  I turned a curve and I was just in time to see him ring the bell and get admitted to the house.& \\
			\cmidrule{2-4}
			\multirow{1}{*}{Orig}& I turned a curve and was just in time to see him ringing the big brass bell, echoing as he was admitted to the house. & \multirow{1}{*}{Entailment} \\
			\cdashlinelr{2-2}
			\multirow{1}{*}{BBA}& I turned a curve and was just in time to see him ringing the big brass bell,  \emph{\red{invoking}} as he was admitted to the \emph{\red{hostels}}. &\multirow{1}{*}{Neutral} \\ 
			\cdashlinelr{2-2}
			\multirow{1}{*}{TF}&  I turned a curve and was just in time to see him \emph{\red{cyclic}} the big brass \emph{\red{chime}}, \emph{\red{noting}} as he was \emph{\red{hospitalised}} to the \emph{\red{sarcophagus}}. & \multirow{1}{*}{Neutral} \\ 
			\midrule
			\multicolumn{2}{l}{Textual Entailment (QNLI)}&Label\\
		\midrule
		Question & What was the main idea of James Hutton's paper?\\
		\cmidrule{2-4}
	
		Orig & In his paper, he explained his theory that the Earth must be much older than had previously been supposed in order to& \multirow{3}{*}{Entailment}\\ & allow  enough time for mountains to be eroded and for sediments to form new rocks at the bottom of the sea, which in turn were\\ & raised up to become dry land.\\
			\cdashlinelr{2-2}
			BBA & In his \emph{\red{journals}}, he explained his theory that the Earth must be much older than had previously been supposed in order to& \multirow{3}{*}{Not Entailment}\\ & allow enough time for mountains to be eroded and for sediments to form new rocks at the bottom of the sea, which in turn were\\ & raised up to become dry land.\\
			\cdashlinelr{2-2}
			TF & In his paper, he explained his theory that the Earth must be much older than had previously been supposed in \emph{\red{writs}} to & \multirow{3}{*}{Not Entailment}\\ & \emph{\red{activation}} enough \emph{\red{timing}} for mountains to be eroded and for sediments to form new rocks at the bottom of the sea, which in turn were \\ &raised up to become dry land.\\
%	\bottomrule
%	\end{tabular}
%\end{adjustbox}
%\begin{adjustbox}{max width=2\columnwidth}
%	\begin{tabular}{ccccc}
%		\toprule
\bottomrule
		\end{tabular}
	\end{adjustbox}
	\label{tab:qual3}
\end{table*}

\end{document}

%% file: fig2.tex
% Figure 5
\begin{figure*}[hbt!]
	\centering
	% Figure 5a
	\begin{subfigure}[t]{0.265\textwidth}
		\begin{tikzpicture}
		\begin{axis}[
		% Figure size
		width=4.5cm,
		height=4.3cm,
		% Plot style
		no marks,
		every axis plot/.append style={thick},
		% Grid
		grid=major,
		% Tick
		scaled ticks = false,
		ylabel near ticks,
		tick pos=left,
		tick label style={font=\small},
		xtick={0, 1000, 2000,3000,4000,5000,6000, 8000},
		xticklabels={0, 1k, 2k, 3k, 4k, 5k, 6k,8k},
		ytick={0, 20, 40, 60, 80, 100},
		yticklabels={0, 20, 40, 60, 80, 100},
		% Label
		label style={font=\small},
		xlabel={Number of queries},
		ylabel={Success rate},
		% Range
		xmin=0,
		xmax=5000,
		ymin=0,
		ymax=110,
        % Legend
		legend style={legend columns=3, at={(1.07, 1.26)}, font=\small},
        ]
		% Ours
		\addplot[red] table [x=qrs, y=ours, col sep=comma]{CSV_final/ours_final/yelp-ours-pwws.csv};
		\addlegendentry{Ours}
		% NES
		%\addplot[green!80!black!100] table [x=qrs, y=LSH, col sep=comma]{CSV_final/wordnet/%BERT_Yelp_5000.csv};
		%\addlegendentry{LSH}
		% Bandits
		\addplot[blue] table [x=qrs, y=PWWS, col sep=comma]{CSV_final/wordnet/BERT_Yelp_5000.csv};
		\addlegendentry{PWWS}
		% PGD
		\draw[dashed] (0, 99.9) -- (10000, 99.9);
		\end{axis}
		\end{tikzpicture}
		%\caption{PWWS}
		%\label{fig:wordnet}
	\end{subfigure}
	\begin{subfigure}[t]{0.235\textwidth}
		\begin{tikzpicture}
		\begin{axis}[
		% Figure size
		width=4.5cm,
		height=4.3cm,
		% Plot style
		no marks,
		every axis plot/.append style={thick},
		% Grid
		grid=major,
		% Tick
		scaled ticks = false,
		ylabel near ticks,
		tick pos=left,
		tick label style={font=\small},
		xtick={0, 1000, 2000,3000,4000,5000,6000, 8000},
		xticklabels={0, 1k, 2k, 3k, 4k, 5k, 6k,8k},
		ytick={0, 20, 40, 60, 80, 100},
		yticklabels={0, 20, 40, 60, 80, 100},
		% Label
		label style={font=\small},
		xlabel={Number of queries},
		ylabel style={at={(-0.2,0.5)}},
		% Range
		xmin=0,
		xmax=2000,
		ymin=0,
		ymax=110,
		% Legend
		legend style={legend columns=3, at={(0.97, 1.26)}, font=\small},
		]
		% Ours
		%\addplot[red] table [x=base, y=Ours, col sep=comma]{data/imagenet_targeted.csv};
		\addplot[red] table [x=qrs, y=ours, col sep=comma]{CSV_final/ours_final/yelp-ours-textfooler.csv};
		\addlegendentry{Ours}
		% NES
		%\addplot[green!80!black!100] table [x=qrs, y=LSH, col sep=comma]{CSV_final/embedding/%BERT_Yelp_5000.csv};
		%\addlegendentry{LSH}
		% Bandtis
		%\addplot[blue] table [x=qrs, y=TF, col sep=comma]{CSV/embedding/BERT_Yelp.csv};
		\addplot[blue] table [x=qrs, y=TF, col sep=comma]{CSV_final/embedding/BERT_Yelp_2000.csv};
		\addlegendentry{TF}
		\draw[dashed] (0, 99.9) -- (10000, 99.9);
		\end{axis}
		\end{tikzpicture}
		%\caption{TF}
		%\caption{ImageNet, targeted}\label{fig:cdf_imagenet_targeted}
	\end{subfigure}
	% Figure 5b
	\begin{subfigure}[t]{0.235\textwidth}
		\begin{tikzpicture}
		\begin{axis}[
		% Figure size
		width=4.5cm,
		height=4.3cm,
		% Plot style
		no marks,
		every axis plot/.append style={thick},
		% Grid
		grid=major,
		% Tick
		scaled ticks = false,
		ylabel near ticks,
		tick pos=left,
		tick label style={font=\small},
		xtick={0, 1000, 2000,3000,4000,5000,6000, 8000},
		xticklabels={0, 1k, 2k, 3k, 4k, 5k, 6k,8k},
		ytick={0, 20, 40, 60, 80, 100},
		yticklabels={0, 20, 40, 60, 80, 100},
		% Label
		label style={font=\small},
		xlabel={Number of queries},
		ylabel style={at={(-0.2,0.5)}},
		% Range
		xmin=0,
		xmax=5000,
		ymin=0,
		ymax=110,
		% Legend
		legend style={legend columns=3, at={(1.005, 1.26), font=\small}},
		]
		% Ours
		\addplot[red] table [x=qrs, y=ours, col sep=comma]{CSV_final/ours_final/yelp-ours-pso.csv};
		\addlegendentry{Ours}
		% NES
		%\addplot[green!80!black!100] table [x=qrs, y=LSH, col sep=comma]{CSV_final/hownet/BERT_Yelp_5000.csv};
		%\addlegendentry{LSH}
		% Bandtis
		%\addplot[blue] table [x=qrs, y=PSO, col sep=comma]{CSV/hownet/BERT_Yelp_2000.csv};
		\addplot[blue] table [x=qrs, y=PSO, col sep=comma]{CSV_final/hownet/BERT_Yelp_5000.csv};
		\addlegendentry{PSO}
		% PGD
		\draw[dashed] (0, 99.9) -- (10000, 99.9);
		\end{axis}
		\end{tikzpicture}
		%\caption{ImageNet, untargeted}\label{fig:cdf_imagenet_untargeted}
		%\caption{PSO}\label{fig:hownet}
	\end{subfigure}
	% Figure 5c
	\begin{subfigure}[t]{0.235\textwidth}
		\begin{tikzpicture}
		\begin{axis}[
		% Figure size
		width=4.5cm,
		height=4.3cm,
		% Plot style
		no marks,
		every axis plot/.append style={thick},
		% Grid
		grid=major,
		% Tick
		scaled ticks = false,
		ylabel near ticks,
		tick pos=left,
		tick label style={font=\small},
		xtick={0, 1000, 2000,3000,4000,5000,6000, 8000},
		xticklabels={0, 1k, 2k, 3k, 4k, 5k, 6k,8k},
		ytick={0, 20, 40, 60, 80, 100},
		yticklabels={0, 20, 40, 60, 80, 100},
		% Label
		label style={font=\small},
		xlabel={Number of queries},
		ylabel style={at={(-0.2,0.5)}},
		% Range
		xmin=0,
		xmax=3000,
		ymin=0,
		ymax=110,
		% Legend
		legend style={legend columns=3, at={(1.005, 1.26)}, font=\small},
		]
		% Ours
		\addplot[red] table [x=qrs, y=ours, col sep=comma]{CSV_final/ours_final/yelp-ours-lsh.csv};
		\addlegendentry{Ours}
		% NES
		%\addplot[green!80!black!100] table [x=qrs, y=LSH, col sep=comma]{CSV_final/embedding/%BERT_Yelp_5000.csv};
		%\addlegendentry{LSH}
		% Bandtis
		%\addplot[blue] table [x=qrs, y=TF, col sep=comma]{CSV/embedding/BERT_Yelp.csv};
		\addplot[blue] table [x=qrs, y=LSH, col sep=comma]{CSV_final/hownet/BERT_Yelp_3000.csv};
		\addlegendentry{LSH}
		\draw[dashed] (0, 99.9) -- (10000, 99.9);
		\end{axis}
		\end{tikzpicture}
		%\caption{LSH}\label{fig:embedding}
		%\caption{ImageNet, targeted}\label{fig:cdf_imagenet_targeted}
	\end{subfigure}
	\caption{The cumulative distribution of the number of queries required for the attack methods against a BERT-base model on the Yelp dataset. We use the HowNet based word substitution when comparing our method against LSH.} %The plots above show that our method finds successful adversarial examples faster than the baseline methods.}
	\label{fig:main}
\end{figure*}

%% file: fig5.tex
% Figure 5
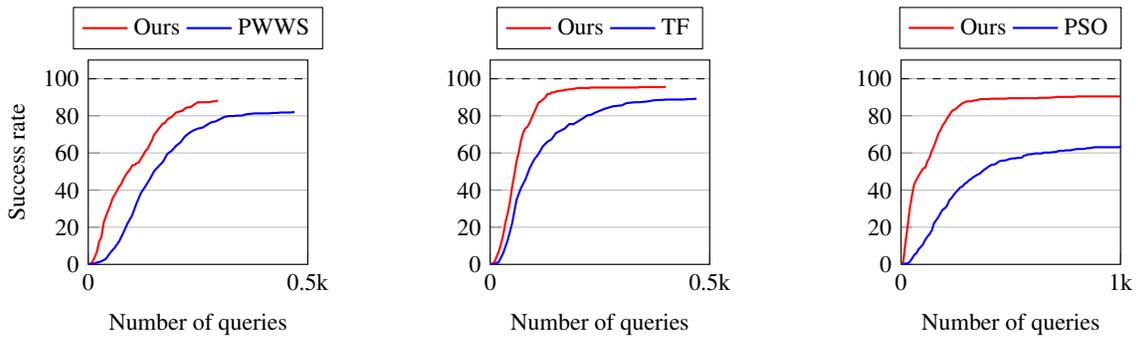
\begin{figure*}[hbt!]
	\centering
	% Figure 5a
	\begin{subfigure}[t]{0.33\textwidth}
	\centering
	\begin{tikzpicture}
		\begin{axis}[
		% Figure size
		width=4.5cm,
		height=4.3cm,
		% Plot style
		no marks,
		every axis plot/.append style={thick},
		% Grid
		grid=major,
		% Tick
		scaled ticks = false,
		ylabel near ticks,
		tick pos=left,
		tick label style={font=\small},
		xtick={0, 500, 1000, 2000,3000,4000,5000,6000, 8000},
		xticklabels={0, 0.5k, 1k, 2k, 3k, 4k, 5k, 6k,8k},
		ytick={0, 20, 40, 60, 80, 100},
		yticklabels={0, 20, 40, 60, 80, 100},
		% Label
		label style={font=\small},
		xlabel={Number of queries},
		ylabel={Success rate},
		% Range
		xmin=0,
		xmax=500,
		ymin=0,
		ymax=110,
        % Legend
		legend style={legend columns=3, at={(1.07, 1.26)}, font=\small},
        ]
		% Ours
		\addplot[red] table [x=qrs, y=ours, col sep=comma]{CSV_final/ours_final/mr-ours-pwws.csv};
		\addlegendentry{Ours}
		% NES
		%\addplot[green!80!black!100] table [x=qrs, y=LSH, col sep=comma]{CSV_final/wordnet/%BERT_Yelp_5000.csv};
		%\addlegendentry{LSH}
		% Bandits
		\addplot[blue] table [x=qrs, y=PWWS, col sep=comma]{CSV_final/wordnet/BERT_MR_5000.csv};
		\addlegendentry{PWWS}
		% PGD
		\draw[dashed] (0, 99.9) -- (10000, 99.9);
		\end{axis}
		\end{tikzpicture}
		%\caption{PWWS}
		%\label{fig:wordnet}
	\end{subfigure}
	\begin{subfigure}[t]{0.31\textwidth}
	\centering
	\begin{tikzpicture}
		\begin{axis}[
		% Figure size
		width=4.5cm,
		height=4.3cm,
		% Plot style
		no marks,
		every axis plot/.append style={thick},
		% Grid
		grid=major,
		% Tick
		scaled ticks = false,
		ylabel near ticks,
		tick pos=left,
		tick label style={font=\small},
		xtick={0, 500, 1000, 2000,3000,4000,5000,6000, 8000},
		xticklabels={0, 0.5k, 1k, 2k, 3k, 4k, 5k, 6k,8k},
		ytick={0, 20, 40, 60, 80, 100},
		yticklabels={0, 20, 40, 60, 80, 100},
		% Label
		label style={font=\small},
		xlabel={Number of queries},
		ylabel style={at={(-0.2,0.5)}},
		% Range
		xmin=0,
		xmax=500,
		ymin=0,
		ymax=110,
		% Legend
		legend style={legend columns=3, at={(0.97, 1.26)}, font=\small},
		]
		% Ours
		%\addplot[red] table [x=base, y=Ours, col sep=comma]{data/imagenet_targeted.csv};
		\addplot[red] table [x=qrs, y=ours, col sep=comma]{CSV_final/ours_final/mr-ours-textfooler.csv};
		\addlegendentry{Ours}
		% NES
		%\addplot[green!80!black!100] table [x=qrs, y=LSH, col sep=comma]{CSV_final/embedding/%BERT_Yelp_5000.csv};
		%\addlegendentry{LSH}
		% Bandtis
		%\addplot[blue] table [x=qrs, y=TF, col sep=comma]{CSV/embedding/BERT_Yelp.csv};
		\addplot[blue] table [x=qrs, y=TF, col sep=comma]{CSV_final/embedding/BERT_MR_5000.csv};
		\addlegendentry{TF}
		\draw[dashed] (0, 99.9) -- (10000, 99.9);
		\end{axis}
		\end{tikzpicture}
		%\caption{TF}
		%\caption{ImageNet, targeted}\label{fig:cdf_imagenet_targeted}
	\end{subfigure}
	% Figure 5b
	\begin{subfigure}[t]{0.31\textwidth}
	\centering
	\begin{tikzpicture}
		\begin{axis}[
		% Figure size
		width=4.5cm,
		height=4.3cm,
		% Plot style
		no marks,
		every axis plot/.append style={thick},
		% Grid
		grid=major,
		% Tick
		scaled ticks = false,
		ylabel near ticks,
		tick pos=left,
		tick label style={font=\small},
		xtick={0, 1000, 2000,3000,4000,5000,6000, 8000},
		xticklabels={0,  1k, 2k, 3k, 4k, 5k, 6k,8k},
		ytick={0, 20, 40, 60, 80, 100},
		yticklabels={0, 20, 40, 60, 80, 100},
		% Label
		label style={font=\small},
		xlabel={Number of queries},
		ylabel style={at={(-0.2,0.5)}},
		% Range
		xmin=0,
		xmax=1000,
		ymin=0,
		ymax=110,
		% Legend
		legend style={legend columns=3, at={(1.005, 1.26), font=\small}},
		]
		% Ours
		\addplot[red] table [x=qrs, y=ours, col sep=comma]{CSV_final/ours_final/mr-ours-pso.csv};
		\addlegendentry{Ours}
		% NES
		%\addplot[green!80!black!100] table [x=qrs, y=LSH, col sep=comma]{CSV_final/hownet/BERT_Yelp_5000.csv};
		%\addlegendentry{LSH}
		% Bandtis
		%\addplot[blue] table [x=qrs, y=PSO, col sep=comma]{CSV/hownet/BERT_Yelp_2000.csv};
		\addplot[blue] table [x=qrs, y=PSO, col sep=comma]{CSV_final/hownet/BERT_MR_5000.csv};
		\addlegendentry{PSO}
		% PGD
		\draw[dashed] (0, 99.9) -- (10000, 99.9);
		\end{axis}
		\end{tikzpicture}
		%\caption{ImageNet, untargeted}\label{fig:cdf_imagenet_untargeted}
		%\caption{PSO}\label{fig:hownet}
	\end{subfigure}
	\caption{The cumulative distribution of the number of queries required for the attack methods against a BERT-base model on the Movie Review dataset.} %The plots above show that our method finds successful adversarial examples faster than the baseline methods.}
	\label{fig:cdmr}
\end{figure*}

%% file: fig3.tex
% Figure 5
\begin{figure*}[hbt!]
	\centering
	% Figure 5a
	\begin{subfigure}[t]{0.265\textwidth}
		\begin{tikzpicture}
		\begin{axis}[
		% Figure size
		width=4.5cm,
		height=4.3cm,
		% Plot style
		no marks,
		every axis plot/.append style={thick},
		% Grid
		grid=major,
		% Tick
		scaled ticks = false,
		ylabel near ticks,
		tick pos=left,
		tick label style={font=\small},
		xtick={0, 1000, 2000,3000,4000,5000,6000, 8000},
		xticklabels={0, 1k, 2k, 3k, 4k, 5k, 6k,8k},
		ytick={0, 20, 40, 60, 80, 100},
		yticklabels={0, 20, 40, 60, 80, 100},
		% Label
		label style={font=\small},
		xlabel={Number of queries},
		ylabel={Success rate},
		% Range
		xmin=0,
		xmax=5000,
		ymin=0,
		ymax=110,
        % Legend
		legend style={legend columns=3, at={(1.07, 1.26)}, font=\small},
        ]
		% Ours
		\addplot[red] table [x=qrs, y=ours, col sep=comma]{CSV_final/ours_final/imdb-ours-pwws.csv};
		\addlegendentry{Ours}
		% NES
		%\addplot[green!80!black!100] table [x=qrs, y=LSH, col sep=comma]{CSV_final/wordnet/%BERT_Yelp_5000.csv};
		%\addlegendentry{LSH}
		% Bandits
		\addplot[blue] table [x=qrs, y=PWWS, col sep=comma]{CSV_final/wordnet/BERT_IMDB_5000.csv};
		\addlegendentry{PWWS}
		% PGD
		\draw[dashed] (0, 99.9) -- (10000, 99.9);
		\end{axis}
		\end{tikzpicture}
		%\caption{PWWS}
		%\label{fig:wordnet}
	\end{subfigure}
	\begin{subfigure}[t]{0.235\textwidth}
		\begin{tikzpicture}
		\begin{axis}[
		% Figure size
		width=4.5cm,
		height=4.3cm,
		% Plot style
		no marks,
		every axis plot/.append style={thick},
		% Grid
		grid=major,
		% Tick
		scaled ticks = false,
		ylabel near ticks,
		tick pos=left,
		tick label style={font=\small},
		xtick={0, 1000, 2000,3000,4000,5000,6000, 8000},
		xticklabels={0, 1k, 2k, 3k, 4k, 5k, 6k,8k},
		ytick={0, 20, 40, 60, 80, 100},
		yticklabels={0, 20, 40, 60, 80, 100},
		% Label
		label style={font=\small},
		xlabel={Number of queries},
		ylabel style={at={(-0.2,0.5)}},
		% Range
		xmin=0,
		xmax=5000,
		ymin=0,
		ymax=110,
		% Legend
		legend style={legend columns=3, at={(0.97, 1.26)}, font=\small},
		]
		% Ours
		%\addplot[red] table [x=base, y=Ours, col sep=comma]{data/imagenet_targeted.csv};
		\addplot[red] table [x=qrs, y=ours, col sep=comma]{CSV_final/ours_final/imdb-ours-textfooler.csv};
		\addlegendentry{Ours}
		% NES
		%\addplot[green!80!black!100] table [x=qrs, y=LSH, col sep=comma]{CSV_final/embedding/%BERT_Yelp_5000.csv};
		%\addlegendentry{LSH}
		% Bandtis
		%\addplot[blue] table [x=qrs, y=TF, col sep=comma]{CSV/embedding/BERT_Yelp.csv};
		\addplot[blue] table [x=qrs, y=TF, col sep=comma]{CSV_final/embedding/BERT_IMDB_5000.csv};
		\addlegendentry{TF}
		\draw[dashed] (0, 99.9) -- (10000, 99.9);
		\end{axis}
		\end{tikzpicture}
		%\caption{TF}
		%\caption{ImageNet, targeted}\label{fig:cdf_imagenet_targeted}
	\end{subfigure}
	% Figure 5b
	\begin{subfigure}[t]{0.235\textwidth}
		\begin{tikzpicture}
		\begin{axis}[
		% Figure size
		width=4.5cm,
		height=4.3cm,
		% Plot style
		no marks,
		every axis plot/.append style={thick},
		% Grid
		grid=major,
		% Tick
		scaled ticks = false,
		ylabel near ticks,
		tick pos=left,
		tick label style={font=\small},
		xtick={0, 1000, 2000,3000,4000,5000,6000, 8000},
		xticklabels={0, 1k, 2k, 3k, 4k, 5k, 6k,8k},
		ytick={0, 20, 40, 60, 80, 100},
		yticklabels={0, 20, 40, 60, 80, 100},
		% Label
		label style={font=\small},
		xlabel={Number of queries},
		ylabel style={at={(-0.2,0.5)}},
		% Range
		xmin=0,
		xmax=5000,
		ymin=0,
		ymax=110,
		% Legend
		legend style={legend columns=3, at={(1.005, 1.26), font=\small}},
		]
		% Ours
		\addplot[red] table [x=qrs, y=ours, col sep=comma]{CSV_final/ours_final/imdb-ours-pso.csv};
		\addlegendentry{Ours}
		% NES
		%\addplot[green!80!black!100] table [x=qrs, y=LSH, col sep=comma]{CSV_final/hownet/BERT_Yelp_5000.csv};
		%\addlegendentry{LSH}
		% Bandtis
		%\addplot[blue] table [x=qrs, y=PSO, col sep=comma]{CSV/hownet/BERT_Yelp_2000.csv};
		\addplot[blue] table [x=qrs, y=PSO, col sep=comma]{CSV_final/hownet/BERT_IMDB_5000.csv};
		\addlegendentry{PSO}
		% PGD
		\draw[dashed] (0, 99.9) -- (10000, 99.9);
		\end{axis}
		\end{tikzpicture}
		%\caption{ImageNet, untargeted}\label{fig:cdf_imagenet_untargeted}
		%\caption{PSO}\label{fig:hownet}
	\end{subfigure}
	% Figure 5c
	\begin{subfigure}[t]{0.235\textwidth}
		\begin{tikzpicture}
		\begin{axis}[
		% Figure size
		width=4.5cm,
		height=4.3cm,
		% Plot style
		no marks,
		every axis plot/.append style={thick},
		% Grid
		grid=major,
		% Tick
		scaled ticks = false,
		ylabel near ticks,
		tick pos=left,
		tick label style={font=\small},
		xtick={0, 1000, 2000,3000,4000,5000,6000, 8000},
		xticklabels={0, 1k, 2k, 3k, 4k, 5k, 6k,8k},
		ytick={0, 20, 40, 60, 80, 100},
		yticklabels={0, 20, 40, 60, 80, 100},
		% Label
		label style={font=\small},
		xlabel={Number of queries},
		ylabel style={at={(-0.2,0.5)}},
		% Range
		xmin=0,
		xmax=5000,
		ymin=0,
		ymax=110,
		% Legend
		legend style={legend columns=3, at={(1.005, 1.26)}, font=\small},
		]
		% Ours
		%\addplot[red] table [x=base, y=Ours, col sep=comma]{data/imagenet_targeted.csv};
		\addplot[red] table [x=qrs, y=ours, col sep=comma]{CSV_final/ours_final/imdb-ours-lsh.csv};
		\addlegendentry{Ours}
		% NES
		%\addplot[green!80!black!100] table [x=qrs, y=LSH, col sep=comma]{CSV_final/embedding/%BERT_Yelp_5000.csv};
		%\addlegendentry{LSH}
		% Bandtis
		%\addplot[blue] table [x=qrs, y=TF, col sep=comma]{CSV/embedding/BERT_Yelp.csv};
		\addplot[blue] table [x=qrs, y=LSH, col sep=comma]{CSV_final/hownet/BERT_IMDB_5000.csv};
		\addlegendentry{LSH}
		\draw[dashed] (0, 99.9) -- (10000, 99.9);
		\end{axis}
		\end{tikzpicture}
		%\caption{LSH}\label{fig:embedding}
		%\caption{ImageNet, targeted}\label{fig:cdf_imagenet_targeted}
	\end{subfigure}
	\caption{The cumulative distribution of the number of queries required for the attack methods against a BERT-base model on the IMDB dataset. We use the HowNet based word substitution when comparing our method against LSH.} %The plots above show that our method finds successful adversarial examples faster than the baseline methods.}
	\label{fig:cdimdb}
\end{figure*}
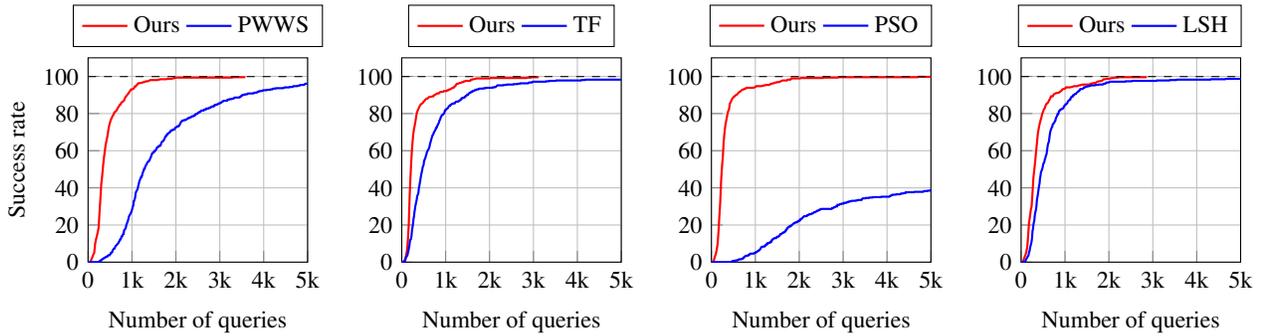

%% file: fig6.tex
% Figure 5
\begin{figure*}[hbt!]
	\centering
	% Figure 5a
	\begin{subfigure}[t]{0.34\textwidth}
	\centering
	\begin{tikzpicture}
		\begin{axis}[
		% Figure size
		width=4.5cm,
		height=4.3cm,
		% Plot style
		no marks,
		every axis plot/.append style={thick},
		% Grid
		grid=major,
		% Tick
		scaled ticks = false,
		ylabel near ticks,
		tick pos=left,
		tick label style={font=\small},
		xtick={0, 400, 1000, 2000,3000,4000,5000,6000, 8000},
		xticklabels={0, 0.4k, 1k, 2k, 3k, 4k, 5k, 6k,8k},
		ytick={0, 20, 40, 60, 80, 100},
		yticklabels={0, 20, 40, 60, 80, 100},
		% Label
		label style={font=\small},
		xlabel={Number of queries},
		ylabel={Success rate},
		% Range
		xmin=0,
		xmax=400,
		ymin=0,
		ymax=110,
        % Legend
		legend style={legend columns=3, at={(1.07, 1.26)}, font=\small},
        ]
		% Ours
		\addplot[red] table [x=qrs, y=ours, col sep=comma]{CSV_final/ours_final/mnli-ours-pwws.csv};
		\addlegendentry{Ours}
		% NES
		%\addplot[green!80!black!100] table [x=qrs, y=LSH, col sep=comma]{CSV_final/wordnet/%BERT_Yelp_5000.csv};
		%\addlegendentry{LSH}
		% Bandits
		\addplot[blue] table [x=qrs, y=PWWS-pre, col sep=comma]{CSV_final/wordnet/BERT_MNLI_5000.csv};
		\addlegendentry{PWWS}
		% PGD
		\draw[dashed] (0, 99.9) -- (10000, 99.9);
		\end{axis}
		\end{tikzpicture}
		%\caption{PWWS}
		%\label{fig:wordnet}
	\end{subfigure}
	\begin{subfigure}[t]{0.31\textwidth}
	\centering
	\begin{tikzpicture}
		\begin{axis}[
		% Figure size
		width=4.5cm,
		height=4.3cm,
		% Plot style
		no marks,
		every axis plot/.append style={thick},
		% Grid
		grid=major,
		% Tick
		scaled ticks = false,
		ylabel near ticks,
		tick pos=left,
		tick label style={font=\small},
		xtick={0, 400, 1000, 2000},
		xticklabels={0, 0.4k, 1k, 2k, 3k, 4k, 5k, 6k,8k},
		ytick={0, 20, 40, 60, 80, 100},
		yticklabels={0, 20, 40, 60, 80, 100},
		% Label
		label style={font=\small},
		xlabel={Number of queries},
		ylabel style={at={(-0.2,0.5)}},
		% Range
		xmin=0,
		xmax=400,
		ymin=0,
		ymax=110,
		% Legend
		legend style={legend columns=3, at={(0.97, 1.26)}, font=\small},
		]
		% Ours
		%\addplot[red] table [x=base, y=Ours, col sep=comma]{data/imagenet_targeted.csv};
		\addplot[red] table [x=qrs, y=ours, col sep=comma]{CSV_final/ours_final/mnli-ours-textfooler.csv};
		\addlegendentry{Ours}
		% NES
		%\addplot[green!80!black!100] table [x=qrs, y=LSH, col sep=comma]{CSV_final/embedding/%BERT_Yelp_5000.csv};
		%\addlegendentry{LSH}
		% Bandtis
		%\addplot[blue] table [x=qrs, y=TF, col sep=comma]{CSV/embedding/BERT_Yelp.csv};
		\addplot[blue] table [x=qrs, y=TF-pre, col sep=comma]{CSV_final/embedding/BERT_MNLI_5000.csv};
		\addlegendentry{TF}
		\draw[dashed] (0, 99.9) -- (10000, 99.9);
		\end{axis}
		\end{tikzpicture}
		%\caption{TF}
		%\caption{ImageNet, targeted}\label{fig:cdf_imagenet_targeted}
	\end{subfigure}
	% Figure 5b
	\begin{subfigure}[t]{0.31\textwidth}
	\centering
	\begin{tikzpicture}
		\begin{axis}[
		% Figure size
		width=4.5cm,
		height=4.3cm,
		% Plot style
		no marks,
		every axis plot/.append style={thick},
		% Grid
		grid=major,
		% Tick
		scaled ticks = false,
		ylabel near ticks,
		tick pos=left,
		tick label style={font=\small},
		xtick={0, 1000, 2000,3000,4000,5000,6000, 8000},
		xticklabels={0,  1k, 2k, 3k, 4k, 5k, 6k,8k},
		ytick={0, 20, 40, 60, 80, 100},
		yticklabels={0, 20, 40, 60, 80, 100},
		% Label
		label style={font=\small},
		xlabel={Number of queries},
		ylabel style={at={(-0.2,0.5)}},
		% Range
		xmin=0,
		xmax=1000,
		ymin=0,
		ymax=110,
		% Legend
		legend style={legend columns=3, at={(1.005, 1.26), font=\small}},
		]
		% Ours
		\addplot[red] table [x=qrs, y=ours, col sep=comma]{CSV_final/ours_final/mnli-ours-pso.csv};
		\addlegendentry{Ours}
		% NES
		%\addplot[green!80!black!100] table [x=qrs, y=LSH, col sep=comma]{CSV_final/hownet/BERT_Yelp_5000.csv};
		%\addlegendentry{LSH}
		% Bandtis
		%\addplot[blue] table [x=qrs, y=PSO, col sep=comma]{CSV/hownet/BERT_Yelp_2000.csv};
		\addplot[blue] table [x=qrs, y=PSO-pre, col sep=comma]{CSV_final/hownet/BERT_MNLI_5000.csv};
		\addlegendentry{PSO}
		% PGD
		\draw[dashed] (0, 99.9) -- (10000, 99.9);
		\end{axis}
		\end{tikzpicture}
		%\caption{ImageNet, untargeted}\label{fig:cdf_imagenet_untargeted}
		%\caption{PSO}\label{fig:hownet}
	\end{subfigure}
		\caption{The cumulative distribution of the number of queries required for the attack methods against a BERT-base model on the MNLI dataset.} %The plots above show that our method finds successful adversarial examples faster than the baseline methods.}
	\label{fig:cdmnli}
\end{figure*}
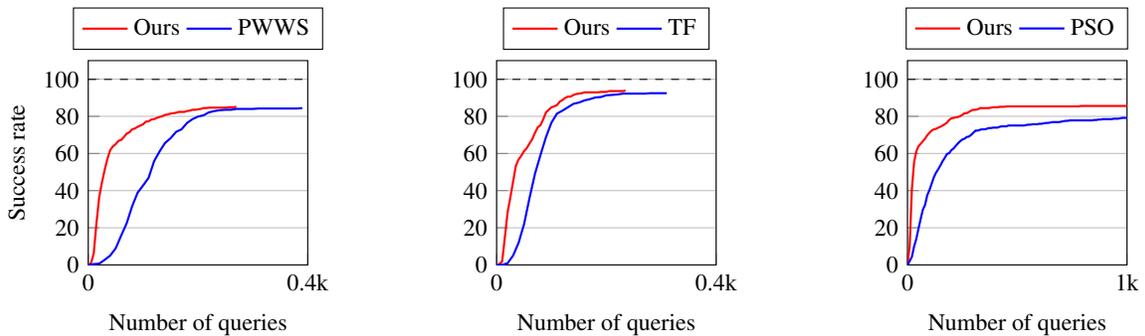